\documentclass[preprint,12pt]{elsarticle}

\usepackage{amssymb}
\usepackage{amsmath}
\usepackage{hyperref}
\usepackage{color}
\usepackage{graphicx}
\usepackage{subcaption}  
\usepackage{xcolor}   
\usepackage{threeparttable}
\usepackage{arydshln}
\usepackage{booktabs}
\usepackage[normalem]{ulem}

\usepackage{pifont}
\newcommand{\xmark}{\ding{55}}

\begin{document}

\begin{frontmatter}

\title{NucFuseRank: Dataset Fusion and Performance Ranking for Nuclei Instance Segmentation}

\author[label1]{Nima Torbati\fnref{equal}}\ead{nima.torbati@dp-uni.ac.at}
\author[label2,label3]{Anastasia Meshcheryakova\fnref{equal}}\ead{anastasia.meshcheryakova@meduniwien.ac.at}
\author[label1]{Ramona Woitek}\ead{Ramona.Woitek@dp-uni.ac.at}
\author[label4,label5]{Sepideh Hatamikia}\ead{Sepideh.Hatamikia@dp-uni.ac.at}
\author[label2,label3]{Diana Mechtcheriakova\fnref{equal2}}\ead{diana.mechtcheriakova@meduniwien.ac.at}
\author[label1]{Amirreza Mahbod\fnref{equal2}}\ead{amirreza.mahbod@dp-uni.ac.at}

\fntext[equal]{Equal contributions}
\fntext[equal2]{Equal contributions}

\affiliation[label1]{organization={Research Center for Medical Image Analysis and Artificial Intelligence, Department of Medicine, Faculty of Medicine and Dentistry, Danube Private University},
            city={Krems an der Donau},
            postcode={3500},
            country={Austria}}

 \affiliation[label2]{organization={Department of Pathophysiology and Allergy Research, Center of Pathophysiology, Infectiology and Immunology, Medical University of Vienna},
             city={Vienna},
             postcode={1090},
             country={Austria}}

 \affiliation[label3]{organization={Comprehensive Center for AI in Medicine (CAIM), Medical University of Vienna},
             city={Vienna},
             postcode={1090},
             country={Austria}}
             
 \affiliation[label4]{organization={Research Center for Clinical AI-Research in Omics and Medical Data Science, Department of Medicine, Faculty of Medicine and Dentistry, Danube Private University},
	city={Krems an der Donau},
	postcode={3500},
	country={Austria}}
	
 \affiliation[label5]{organization={Austrian Center for Medical Innovation and Technology},
	city={Wiener Neustadt},
	postcode={2700},
	country={Austria}}
\date{}

\begin{abstract}
Nuclei instance segmentation in hematoxylin and eosin (H\&E)-stained images plays an important role in automated histological image analysis, with various applications in downstream tasks. While several machine learning and deep learning approaches have been proposed for nuclei instance segmentation, most research in this field focuses on developing new segmentation algorithms and benchmarking them on a limited number of arbitrarily selected public datasets.

In this work, rather than focusing on model development, we focused on the datasets used for this task. Based on an extensive literature review, we identified manually annotated, publicly available datasets of H\&E-stained images for nuclei instance segmentation and standardized them into a unified input and annotation format. Using two state-of-the-art segmentation models, one based on convolutional neural networks (CNNs) and one based on a hybrid CNN and vision transformer architecture, we systematically evaluated and ranked these datasets based on their nuclei instance segmentation performance. Furthermore, we proposed a unified test set (NucFuse-test) for fair cross-dataset evaluation and a unified training set (NucFuse-train) for improved segmentation performance by merging images from multiple datasets.

By evaluating and ranking the datasets, performing comprehensive analyses, generating fused datasets, conducting external validation, and making our implementation publicly available, we provided a new benchmark for training, testing, and evaluating nuclei instance segmentation models on H\&E-stained histological images.
\end{abstract}




\end{frontmatter}

\section{Introduction}
\label{sec:intro}
Analyzing histological whole-slide images (WSIs) is still considered the gold standard method for the diagnosis and prognosis of various diseases, including different types of cancer~\cite{jmp7010002}. To visualize cell and tissue structures, various staining types are commonly used. Among these, hematoxylin and eosin (H\&E) staining is the most widely applied~\cite{10463355}. When particular protein markers need to be visualized, other staining techniques, such as immunohistochemistry, can also be applied~\cite{MEBRATIE2024154}.

Traditionally, histological image assessment relies on conventional light microscopy, with pathologists examining tissue sections through microscope optics.
By systematically navigating across regions of interest at multiple magnifications and applying their domain expertise, pathologists derive diagnoses~\cite{doi:10.5946/ce.2023.036}.
With the advent of digital slide scanners, which enable scanning of histological slides into digital WSIs that can be stored on computers, numerous semi- and fully-automatic approaches have emerged for histological image analysis~\cite{KOMURA2025383}, including tasks such as patch-based and whole slide-based classification~\cite{mahbod2018breast, GHAFFARILALEH2022102474}, segmentation of diverse anatomical parts of the tissues within the images~\cite{torbati2025acs, torbati2025multi}, and histological image retrieval~\cite{WANG2023102645, saeidi2024leveraging}. Among these tasks, nuclei instance segmentation is considered a fundamental step, as it provides relevant and essential information for downstream analysis tasks~\cite{monuseg}. For instance, in breast cancer grading, nuclear density and nuclear counts are important features that can be derived when accurate nuclei instance segmentation masks are available~\cite{DUFER1993131, Das2020}.

Regarding methodological approaches, various techniques and models have been proposed, ranging from conventional image processing methods to advanced machine learning- and deep learning (DL)-based approaches~\cite{belsare2012histopathological, Hossain2024}.

For nuclei instance segmentation, DL-based methods proposed in the literature can generally be categorized into two main groups~\cite{NUNES2025103360, 10.3389/fmed.2022.978146}: (i) detection-based methods, such as Mask R-CNN and its enhanced variants~\cite{bioengineering11100994, pmlr-v156-bancher21a}, and (ii) segmentation-based approaches, including multi-encoder and multi-decoder convolutional neural network (CNN) architectures as well as vision transformer (ViT)-based or hybrid CNN–ViT approaches. Models such as HoVerNet~\cite{graham2019hover}, HoVerNeXt~\cite{baumann2024hover}, DDU-Net~\cite{10.3389/fmed.2022.978146}, Stardist~\cite{9854534} and CellViT~\cite{HORST2024103143} are examples of state-of-the-art DL-based approaches for nuclei instance segmentation and classification tasks. To evaluate these models, a number of publicly available datasets, such as MoNuSeg~\cite{monuseg} or NuInsSeg~\cite{mahbod2024nuinsseg} datasets, exist, which are commonly used as benchmarks.

Although many methods have demonstrated excellent segmentation or detection performance, most research in this field has focused on developing or enhancing new models. These models are typically evaluated on a limited number of publicly available datasets, often selected arbitrarily from the larger pool of datasets available for nuclei instance segmentation. Moreover, the formats of input images and corresponding segmentation masks vary across datasets, which makes it difficult to seamlessly conduct straightforward and fair experimental comparisons when evaluating model performance.

In this work, instead of developing new models for nuclei instance segmentation, we aimed to provide a comprehensive benchmark and analysis of publicly available datasets used for training and evaluation. We performed a thorough literature search to identify all manually annotated, publicly available datasets for nuclei instance segmentation, which resulted in the selection of 10 datasets. We standardized the input image formats and corresponding segmentation mask formats across all datasets to enable straightforward experimentation without additional dataset-specific preprocessing, thereby ensuring a fair and consistent evaluation pipeline. Based on predefined criteria, we selected a fixed number of images from each dataset to form a universal test set, referred to as NucFuse-test. The remaining images from each dataset were used for training. Using two state-of-the-art nuclei instance segmentation models, one based on CNN and one based on a hybrid CNN–ViT architecture, we conducted a series of experiments by training each model on the training images of individual datasets and evaluating their performance on the unified test set. This allowed us to rank the datasets based on their generalization performance on NucFuse-test data. We merged the training images from multiple datasets to construct a combined training set, referred to as NucFuse-train, and investigated model performances when trained on this merged dataset.

In summary, the main contributions of this study are as follows:

\begin{itemize}
\item We standardize the formats of publicly available nuclei instance segmentation datasets, including both input images and segmentation masks, enabling seamless and reproducible experimentation.

\item We provide a comprehensive benchmark and ranking of manually annotated publicly available datasets for the nuclei instance segmentation task.

\item We introduce a unified fused test dataset (NucFuse-test) for fair cross-dataset evaluation.

\item We construct a fused training dataset (NucFuse-train) consisting of training images from multiple publicly available datasets considered in this study.

\item We make all code and all created datasets, including NucFuse-test and NucFuse-train, publicly available for future research and benchmarking. 

\end{itemize}

\begin{figure}[ht]
    \centering
    \includegraphics[width=\textwidth]{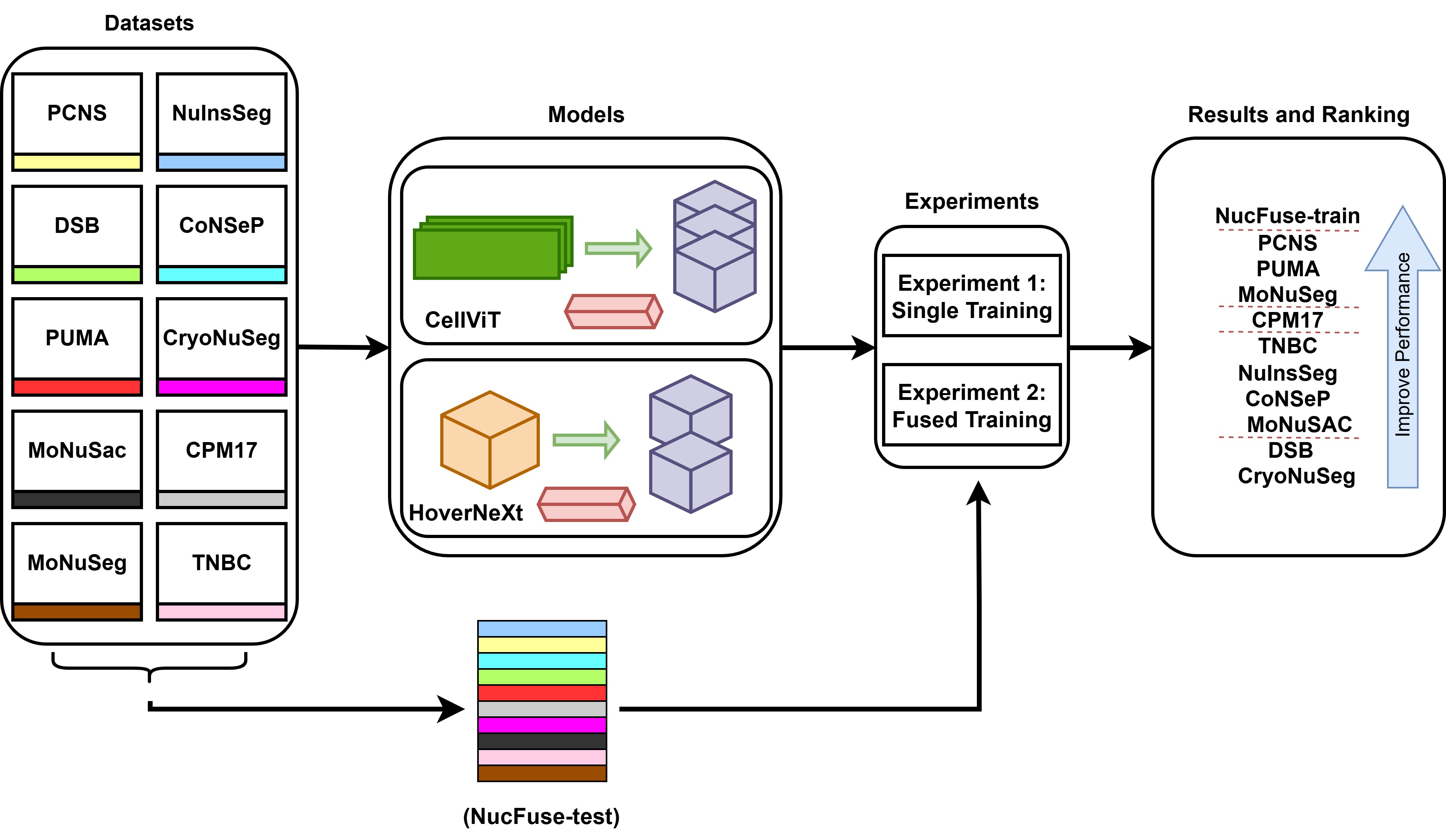}
    \footnotesize  \caption{
    An overview of the entire workflow. First, a unified test set (NucFuse-test) was constructed from all datasets to evaluate the experiments. The remaining images were then used to train two state-of-the-art models, CellViT~\cite{HORST2024103143} and HoVerNeXt~\cite{baumann2024hover}, under two experimental setups: single-dataset training (e.g., training only on PCNS or only on NuInsSeg) and fused-dataset training (progressively merging the training datasets). Based on the results of Experiment 1, the datasets were ranked according to their performance on the unified test set. Based on the results of Experiment 2, an optimal merged training dataset (NucFuse-train) was introduced.}
    \label{fig:overal}
\end{figure}

\section{Methods}
\label{sec:me}
An overview of the entire pipeline is shown in Figure~\ref{fig:overal}. The overall goal was to analyze and rank publicly available datasets for the nuclei instance segmentation task based on the performance of models trained on them and evaluated on a unified test set, and to introduce a new merged training dataset. The results of this study can serve as a benchmark for dataset selection for model training and for model evaluation in future research.

To achieve these goals, we first constructed a unified test set (NucFuse-test) by selecting a fixed number of images from each dataset based on predefined criteria. The remaining images were then used to train two state-of-the-art models, CellViT~\cite{HORST2024103143} and HoVerNeXt~\cite{baumann2024hover}, under two experimental setups: single-dataset training and fused-dataset training. The experimental results from single-dataset training were used to rank the datasets and results from fused-dataset training were used to form an optimal merged training dataset (NucFuse-train). 

In the following subsection, the details of our approach are described.

\subsection{Datasets}
\label{subsec1}

In this work, we compiled publicly available, manually annotated datasets for the task of nuclei instance segmentation. In addition to datasets originally designed for this task, such as MoNuSeg~\cite{monuseg} and CryoNuSeg~\cite{mahbod2021cryonuseg_org}, we also included nuclei instance segmentation and classification datasets, such as MoNuSAC~\cite{9446924} and CoNSeP~\cite{graham2019hover}. For the latter datasets, all classes were merged into a single foreground class to make them suitable for the nuclei instance segmentation task. This results in a total of 10 datasets, of which seven were originally designed for nuclei instance segmentation, and three were derived from nuclei instance segmentation and classification datasets.

The main statistics of these datasets are summarized in Table~\ref{tab:datasets} (No. 1 to No. 10). For each dataset, Table~\ref{tab:datasets} reports the total number of image tiles, the total number of annotated nuclei, the number of organs from which the images of tissue sections were obtained, and the tile size (if multiple sizes are present, the range of tile sizes). Additionally, for datasets that include nuclei classes, this information is indicated in the classification column of the table. Furthermore, the table includes the nuclei density,  defined as the mean number of annotated nuclei per $256\times256$ patch for each dataset. For example, the CryoNuSeg dataset contains an average of 63.26 annotated nuclei per $256\times256$ patch (the highest density among all datasets), while the NuInsSeg dataset has a nuclei density of 11.54, which is the lowest among all datasets.

These 10 selected datasets were chosen from a list of 24 publicly available datasets for nuclei instance segmentation or detection in histology images. The complete list of excluded datasets is reported in Table S1 of the supplementary materials, together with the corresponding main reason for exclusion. 

Since our focus was on the instance segmentation task, we excluded datasets designed only for detection, such as OCELOT~\cite{Ryu_2023_CVPR}, as these datasets provide annotations only for nuclei centers. Additionally, we excluded datasets that are subsets of larger datasets; Kumar~\cite{kumar2017dataset} and CPM-15~\cite{vu2019methods} fall into this category. We also excluded weakly annotated datasets, such as Janowczyk~\cite{janowczyk2016deep}. In these datasets, either only parts of the images are annotated or only specific nuclei classes are annotated. Furthermore, we excluded datasets such as NuCLS~\cite{liu2022panoptic} and Lizard~\cite{graham2021lizard}, which rely on semi-automatic annotations. We did not include these datasets mainly for two reasons. First, semi-automatic pipelines typically employ a DL backbone to generate initial annotations, which may introduce implicit bias toward the characteristics of the model used for annotation. Second, the backbone models are often pretrained using other publicly available datasets; therefore, including such semi-automatically annotated datasets in our experimental setup could introduce unintended dependencies and potential data leakage, compromising the fairness of the evaluation. However, among the semi-automatically annotated datasets, we retained PanNuke~\cite{gamper2020pannuke} dataset (11th dataset in Table~\ref{tab:datasets}) for comparison purposes. Finally, we excluded datasets such as UCSB~\cite{gelasca2008evaluation} for which either the input images or the corresponding segmentation masks are no longer available.

\subsection{Merged Test and Train Sets}
To validate the performance of the studied models, we constructed a unified test set (NucFuse-test) using samples from all datasets considered in this study. Specifically, 14 image tiles were selected from each dataset. If a dataset originally included a test set or a validation set, the 14 tiles were selected from these subsets, with priority given to the test set. If a dataset did not provide either a test or a validation set, the 14 tiles were randomly sampled from the entire dataset. We chose the value of 14 because MoNuSeg~\cite{monuseg} has the smallest test set among the datasets with an official test split, containing 14 images. We also set a minimum tile-size threshold of $256\times256$ pixels when sampling the test set. For the MoNuSAC dataset, images containing macrophages were excluded to reduce bias toward this dataset, as macrophages were only included in MoNuSAC and their annotations/segmentations were generally less reliable~\cite{10.1145/3632047.3632052, JIMENEZ2024108586}. The resulting NucFuse-test set reduces the influence of datasets with larger sample sizes while preserving the original structure of each dataset. After selecting 14 image tiles per dataset, all remaining tiles were used for training. By progressively merging training tiles across datasets, we also created a merged training dataset (NucFuse-train) to investigate whether combining publicly available data could improve nuclei instance segmentation performance.

We make all generated datasets publicly available in a FigShare repository at \url{https://figshare.com/s/c81d0c3401e1bea007f6}. The repository contains three main folders. The first folder, named \texttt{original\_split}, contains all datasets in their original form (original train/test split, or the entire dataset if no official split exists), using an identical input image format (\texttt{.tif}) and segmentation mask format (\texttt{.npy}). The second folder, named \texttt{custom\_split}, contains all datasets using the input image and segmentation mask formats based on our custom split. Finally, the third folder, named \texttt{merged\_split}, contains NucFuse-test and NucFuse-train.

\begin{table}[h!]
\centering
\resizebox{\textwidth}{!}{
\begin{tabular}{c l c c c c c c c}
\hline
\textbf{No.} & \textbf{Name} & \textbf{Tiles} & \textbf{Nuclei} & \textbf{Organs} & \textbf{TS} & \textbf{CL} & \textbf{ND} & \textbf{TT}\\
\hline
1 & PCNS \cite{Hou2019Pan} & 1,356 & 37,313 & 9 & 256$\times$256 & \xmark  & 27.51 & 1,342\\
2 & PUMA \cite{schuiveling2025novel} & 205 & 97,187 & 8 & 1024$\times$1024 & $\checkmark$  & 29.63 & 3,056\\
3 & MoNuSeg \cite{monuseg} & 51 & 30,827 & 9 & 1000$\times$1000 & \xmark & 39.61 & 592\\
4 & CPM17 \cite{vu2019methods} & 64 & 7,570 & 4 & 500$\times$500 -- 600$\times$600 & \xmark & 27.25 & 285\\
5 & TNBC \cite{naylor2018segmentation} & 50 & 4,056 & 1 & 512$\times$512 & \xmark & 20.28 & 144\\
6 & NuInsSeg \cite{mahbod2024nuinsseg} & 665 & 30,698 & 31 & 512$\times$512 & \xmark  & 11.54 & 2,604\\
7 & CoNSeP \cite{graham2019hover} & 41 & 24332 & 41 & 1000$\times$1000 & $\checkmark$  & 38.89 & 432\\
8 & MoNuSAC \cite{9446924} & 310 & 46,828 & 4 & 33$\times$86 -- 1771$\times$1760 & $\checkmark$  & 21.31 & 2,692\\
9 & DSB \cite{caicedo2019nucleus} & 107 & 4,529 & na & 256$\times$320 & \xmark  & 33.86& 186\\
10 & CryoNuSeg \cite{mahbod2021cryonuseg_org} & 30 & 7,592 & 10 & 512$\times$512 & \xmark  & 63.26 & 64\\
\hline
11 & PanNuke \cite{gamper2020pannuke} & 7,904 & 189,744 & 19 & 256$\times$256 & $\checkmark$  & 24.00 & 6,320\\
\hline
12 & NucFuse-train & 2,739 & 248,815 & na &33$\times$86 -- 1771$\times$1760 & \xmark & 24.25 & 11,333\\
13 & NucFuse-test & 140 & 34,525 & na & 256$\times$256 -- 1400$\times$1872 & \xmark & 32.27 & 140\\
\hline
\end{tabular}
}
\caption{Summary of the datasets used in this study. Datasets listed in the table (No. 1 to No. 10) are manually annotated for the task of nuclei instance segmentation. The PanNuke dataset (No. 11) is a semi-automatically generated dataset, and NucFuse-train and NucFuse-test are merged datasets. TS: tile size; CL: classification label; cross mark indicates dataset originally designed for nuclei instance segmentation and check mark indicates dataset designed for nuclei instance segmentation and classification; ND: average nuclei density in a $256\times256$ tile; TT: number of $256\times256$ training image tiles, with one exception for NucFuse-test, where this value indicates the total number of test images with varying sizes, na: not available}
\label{tab:datasets}
\end{table}

\subsection{Models}
To evaluate and compare instance segmentation performance across the datasets, we employed two state-of-the-art segmentation models: HoVerNeXt and CellViT. HoVerNeXt is among the latest CNN-based architectures, while CellViT belongs to the ViT family. Both models are originally capable of performing nuclei classification; however, this functionality was disregarded in our experiments, as we focused solely on the nuclei instance segmentation task in this study.

\subsubsection{HoVerNeXt}
In an encoder–decoder architecture, HoVerNeXt integrates ConvNeXt-V2~\cite{woo2023convnext} encoders to enhance nuclei segmentation accuracy in histological images. Its decoder consists of two parallel branches: a class decoder and an instance decoder. In our experiments, only the instance decoder was used. HoVerNeXt is also much faster than its predecessor, HoVerNet~\cite{graham2019hover}, while providing comparable results.

\subsubsection{CellViT}
CellViT is an encoder–decoder-based model featuring a single encoder and three decoder branches. The novelty of CellViT lies in its use of a ViT~\cite{dosovitskiy2020image} as the encoder backbone. To achieve optimal performance, the model leverages ViT encoders pretrained on histological image data. Similar to HoVerNet, the decoder consists of three branches responsible for binary segmentation, class prediction (disregarded in our experiments), and distance map estimation. Equipped with its ViT encoder, CellViT achieved superior performance compared to many other models, including HoVerNet.

\section{Experiments}
This section presents the experimental setup and evaluation details. We conducted two main experiments, as described below.

\subsection{Experiment 1: Single-Dataset Training}
\label{sec:experiment1}
In this set of experiments, we trained independent HoVerNeXt and CellViT models using the training data of each individual dataset listed in Table~\ref{tab:datasets} (No. 1 to No. 10). To ensure a uniform training scheme, we first generated non-overlapping $256\times256$ pixel patches from all images in each dataset. If the size of an image was smaller than 256 pixels in either dimension, we applied white-padding to increase its size to $256\times256$ pixels. The number of $256\times256$ patches used for training each dataset is reported in the last column of Table~\ref{tab:datasets}. We used an 80\%/20\% split for training and validation, respectively. All trained models were evaluated on the generated unified test set (NucFuse-test) and were subsequently ranked based on their nuclei instance segmentation performance.

\subsection{Experiment 2: \textit{K}-Best Dataset Training}
\label{sec:experiment2}
Using the rankings obtained from Experiment set 1, we constructed fused datasets by combining the top-\textit{k} best datasets to identify the optimal joint training set. For this purpose, we trained both models (HoVerNeXt and CellViT) on fused datasets for $\textit{k}=1$ to $\textit{k}=10$, where $\textit{k}=9$ represents the NucFuse-train dataset. The rationale for choosing $\textit{k}=9$ for NucFuse-train is provided in Section~\ref{sec:K-Best Dataset Training}. Similar to Experiment set 1, we evaluated the performance of all models using the unified test set (NucFuse-test).

\subsection{Additional Experiments}
While the main focus of this study is on Experiment 1 and Experiment 2, we also performed additional experiments. These experiments investigated the effect of stain normalization on performance, compared results with training with a semi-automatically generated dataset (PanNuke~\cite{10.1007/978-3-030-23937-4_2}), and compared results with a recently developed foundation model for nuclei and cell segmentation (CellSAM~\cite{israel2025cellsam}).

\subsection{Evaluation}
For evaluation, we employed widely used metrics including panoptic quality (PQ), aggregated Jaccard index (AJI), Dice score, precision, and recall. Among these metrics, PQ is the most commonly used index for benchmarking~\cite{10.3389/fmed.2022.978146, graham2019hover, 9446924}; therefore, we used it as the main evaluation metric for dataset ranking in our experiments. The details of these indices are described below:

\begin{itemize}
    \item \textbf{Panoptic quality (PQ):}
    \[
    PQ = \frac{\sum_{(p,g) \in TP} \mathrm{IoU}(p,g)}{|TP| + \frac{1}{2}|FP| + \frac{1}{2}|FN|}
    \]
    where \(TP\) denotes predicted instances that can be uniquely paired with a ground truth instance with \(\mathrm{IoU} > 0.5\); instances that cannot be paired are counted as \(FP\). \(FN\) represents ground truth instances that are not paired with any prediction. \(\mathrm{IoU}\) is the intersection over union of a paired ground truth and predicted instance.
    \item \textbf{Aggregated Jaccard index (AJI):}
    \[
    AJI = \frac{\sum_{i} |G_i \cap P_i|}{\sum_{i} |G_i \cup P_i| + \sum_{k} |P_k| + \sum_{l} |G_l|}
    \]
    where \(G_i\) and \(P_i\) are matched ground truth and predicted instances, and \(P_k\) and \(G_l\) are unmatched predicted and ground truth instances, respectively.
    \item \textbf{Dice score:}
    \[
    \mathrm{Dice} = \frac{2 \sum_{i} |G_i \cap P_i|}{\sum_{} |G| + \sum_{} |P|}
    \]
    \item \textbf{Precision:}
    \[
    \mathrm{Precision} = \frac{TP}{TP + FP}
    \]
    \item \textbf{Recall:}
    \[
    \mathrm{Recall} = \frac{TP}{TP + FN}
    \]
\end{itemize}

For a better comparison between datasets, we also defined a dataset-based cross-correlation matrix as follows:
\[
C_{i,j} = \frac{P(D_i, D_j) + P(D_j, D_j)}{2\, P(D_j, D_j)}
\]

where $C_{i,j}$ denotes the cross-correlation between datasets $D_i$ and $D_j$, and $P(D_i, D_j)$ represents the model performance on the test set derived from dataset $D_j$ when the model is trained on dataset $D_i$.

\subsection{Implementation Details}
All experiments were conducted on a workstation equipped with an AMD Ryzen 9 7950X CPU, 96 GB of RAM, and an NVIDIA RTX 4090 GPU. For both models, we adhered to the original implementations for training and inference. Specifically, we used \textit{convnextv2tiny} encoder for HoVerNeXt and the \textit{ViT256} encoder for CellViT. All models were trained with a batch size of 5 for 300 epochs. Since both models output horizontal and vertical distance maps, these maps were used in the post-processing step for nuclie instance segmentation. Our full training and evaluation implementation is available in our GitHub repository:\url{https://github.com/masih4/NucleiAnalysis}.

\begin{table}[h!]
\centering
\resizebox{0.9\textwidth}{!}{
\begin{tabular}{l c c c c c c c c}
\hline
  \textbf{Dataset} & \textbf{Model} & \textbf{PQ} & \textbf{AJI} & \textbf{Dice} & \textbf{Precision} & \textbf{Recall} & \textbf{PQ Rank} & \textbf{Mean Rank}\\
\hline
PCNS \cite{Hou2019Pan} & HoVerNeXt & 55.66 & 58.31 & 68.81 & 72.64 & 76.82&  2&  1\\
  & CellViT  & 55.14 & 57.60 & 68.20 & 73.21 & 74.67  &1  &\\ \\
PUMA \cite{schuiveling2025novel} & HoVerNeXt & 55.75 & 57.83 & 71.57 & 78.62 & 70.83 & 1 & 2\\
   & CellViT & 54.74 & 57.10 & 72.05 & 77.89 & 69.62 &   3 &\\ \\
 MoNuSeg \cite{monuseg} & HoVerNeXt & 53.38 & 56.08 & 68.45 & 71.05 & 73.72&  3& 3\\ 
   & CellViT  & 55.11 & 57.55 & 69.43 & 75.09 & 72.26 & 2 &\\ \\
\hdashline

 CPM17 \cite{vu2019methods} & HoVerNeXt & 51.86 & 54.49 & 65.38 & 68.05 & 71.81 & 4 & 4\\
   & CellViT  & 51.58 & 54.36 & 66.18 & 70.36 & 70.01 &  4 &\\ \\
\hdashline

 TNBC \cite{naylor2018segmentation} & HoVerNeXt & 46.39 & 48.20 & 65.94 & 65.82 & 62.44 &  7& 5\\
   & CellViT  & 48.67 & 52.76 & 64.85 & 65.13 & 68.85 &  5\\ \\
 NuInsSeg \cite{mahbod2024nuinsseg} & HoVerNeXt & 44.92 & 44.94 & 68.04 & 77.01 & 55.66 & 8 & 6\\
   & CellViT  & 48.54 & 50.38 & 66.06 & 71.77 & 63.35 &  6 &\\ \\
CoNSeP \cite{graham2019hover} & HoVerNeXt & 46.41 & 50.64 & 63.44 & 61.02 & 68.24 & 6 & 7\\
   & CellViT  & 44.97 & 50.20 & 62.49 & 57.05 & 67.51 &  7 &\\ \\
MoNuSAC \cite{9446924} & HoVerNeXt & 47.28 & 47.28 & 65.73 & 77.70 & 55.34 &  5& 8\\
   & CellViT  & 40.40 & 40.40 & 61.39 & 71.44 & 45.84 &  9 &\\ \\
\hdashline

DSB \cite{caicedo2019nucleus} & HoVerNeXt & 43.83 & 45.99 & 60.13 & 60.57 & 61.97 &  9& 9\\
   & CellViT  & 38.44 & 44.24 & 56.53 & 49.26 & 59.72 &  10 &\\ \\
 CryoNuSeg \cite{mahbod2021cryonuseg_org} & HoVerNeXt & 35.57 & 36.57 & 60.62 & 60.31 & 48.16 &  10& 10\\
   & CellViT  & 40.82 & 45.49 & 60.08 & 55.90 & 59.91&  8 &\\ 
\hline
  Mean & HoVerNeXt & 48.10 & 50.03 & 65.81 & 69.27 & 64.49 & & \\
   (across datasets)& CellViT & 47.84 & 51.00 & 64.72 & 66.71 & 65.17 &  &\\ 
\hline

\end{tabular}
}
\caption{Ranking of datasets based on their mean panoptic quality (PQ) performance (\%) on the NucFuse-test set. In addition to PQ, results for other metrics (aggregate Jaccard index (AJI) (\%), Dice score (\%), precision (\%), and recall (\%)) are also reported for both models (HoVerNeXt and CellViT).
}
\label{tab:single_res}
\end{table}

\section{Results \& Discussion}
\label{sec:res}
The experimental results and discussion from Experiment 1, Experiment 2, and additional experiments are presented in this section.
\subsection{Experiment 1: Single-Dataset Training}
The results of Experiment 1, described in  Section~\ref{sec:experiment1}, in terms of PQ, AJI, Dice score, precision, and recall, are shown in Table~\ref{tab:single_res}. The datasets were ordered and ranked based on their mean PQ scores on the NucFuse-test set. While the last column shows the exact ranking of the datasets, we also categorized them into four groups (separated by dashed lines in the table). PCNS, PUMA, and MoNuSeg form the first group and can be considered the three best-performing datasets for both HoVerNeXt and CellViT, with a large margin compared to the other datasets. CPM17 is in the second group; although inferior to the datasets in the first group, it still delivers substantially better average performance than the lower-ranked datasets. TNBC, NuInsSeg, CoNSeP, and MoNuSAC are in the third group, and finally DSB and CryoNuSeg are in the fourth group, showing the lowest average performance. Of note, the WSIs used to create the CryoNuSeg dataset are based on frozen-sectioned tissue samples which, due to different preparation techniques, may result in different image quality. This may limit generalization and lead to inferior performance compared to other datasets, which are mainly based on standard formalin-fixed paraffin-embedded (FFPE) samples~\cite{mahbod2021cryonuseg_org}.

Furthermore, as the ranking indicates, some datasets with smaller training sample sizes can still deliver better performance than datasets with larger training sets. For example, CPM17 and TNBC, with only 285 and 144 training samples, respectively, achieved better overall results compared to much larger datasets such as NuInsSeg and MoNuSAC, which contain 2,604 and 2,692 training samples, respectively. This finding confirms that training set size is not the only contributing factor to improved performance, and that other criteria, such as annotation quality and diversity in the training images, influence the results.

Comparing the models, HoVerNeXt and CellViT achieved comparable performance, with mean PQ scores of 48.10\% for HoVerNeXt and 47.84\% for CellViT, with HoVerNeXt performing slightly better. CellViT’s performance could likely be improved by using a larger encoder (e.g., incorporating a pretrained foundation model as the encoder, similar to CellViT++~\cite{HORST2026109206}); however, achieving the highest possible performance was not the primary objective of this study. In terms of model size, HoVerNeXt contained approximately 36 million trainable parameters, compared with about 46 million trainable parameters for CellViT. Therefore, HoVerNeXt may be preferable when computational resources are limited.

It is also worth notifying that for the higher-ranked datasets, particularly those in group 1 and group 2, the performance differences between HoVerNeXt and CellViT were smaller than for the lower-ranked datasets. For instance, for the PCNS and PUMA datasets, the PQ differences between the two models were 0.42\% and 1.01\%, respectively, whereas for DSB and CryoNuSeg, the performance differences were much larger (5.39\% and 5.25\%, respectively).

\begin{figure}[]
    \centering
    \begin{subfigure}[b]{0.43\textwidth}
        \centering
        \includegraphics[height=3.5cm]{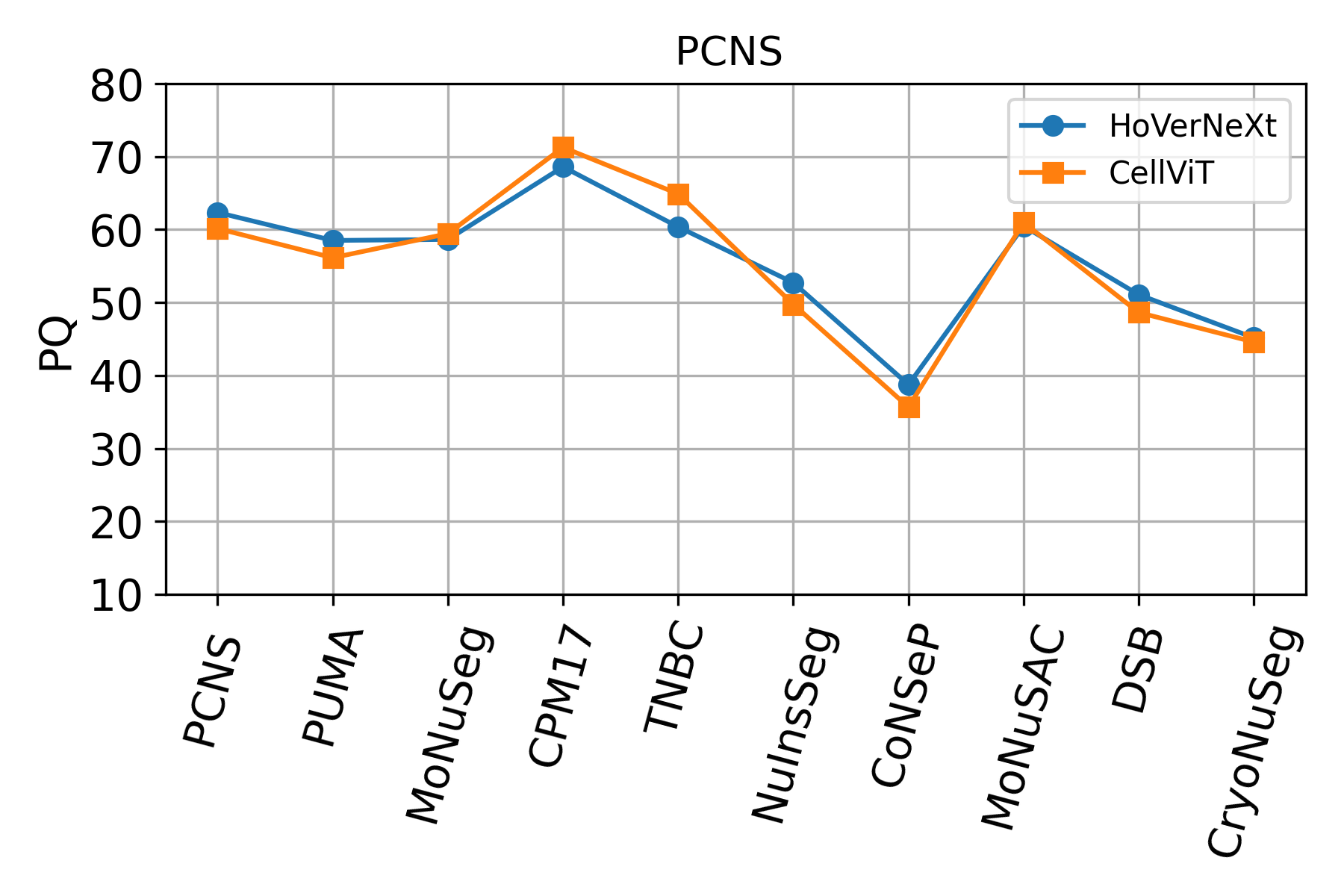}
    \end{subfigure}
    \begin{subfigure}[b]{0.43\textwidth}
        \centering
        \includegraphics[height=3.5cm]{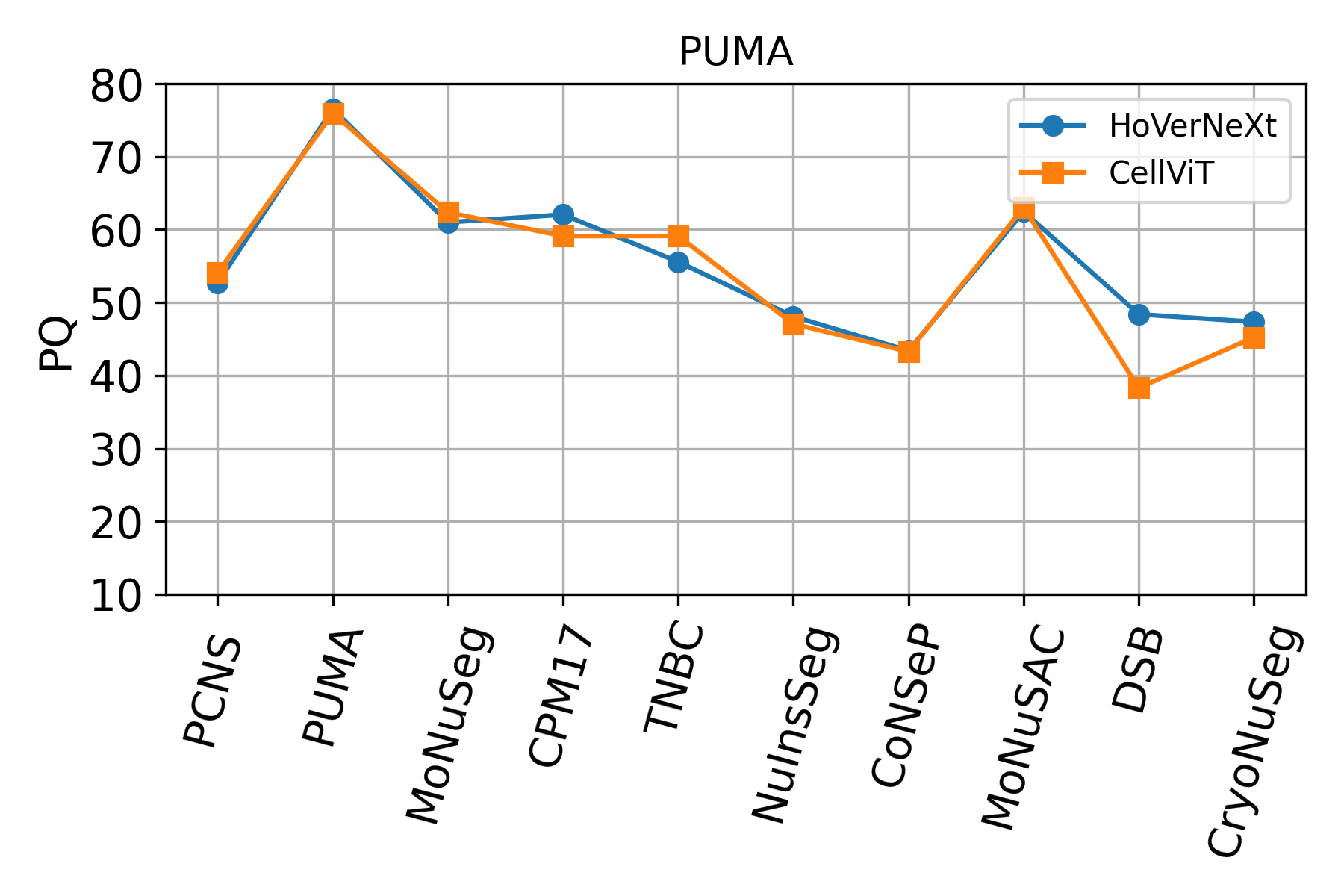}
    \end{subfigure} \\
    \begin{subfigure}[b]{0.43\textwidth}
        \centering
        \includegraphics[height=3.5cm]{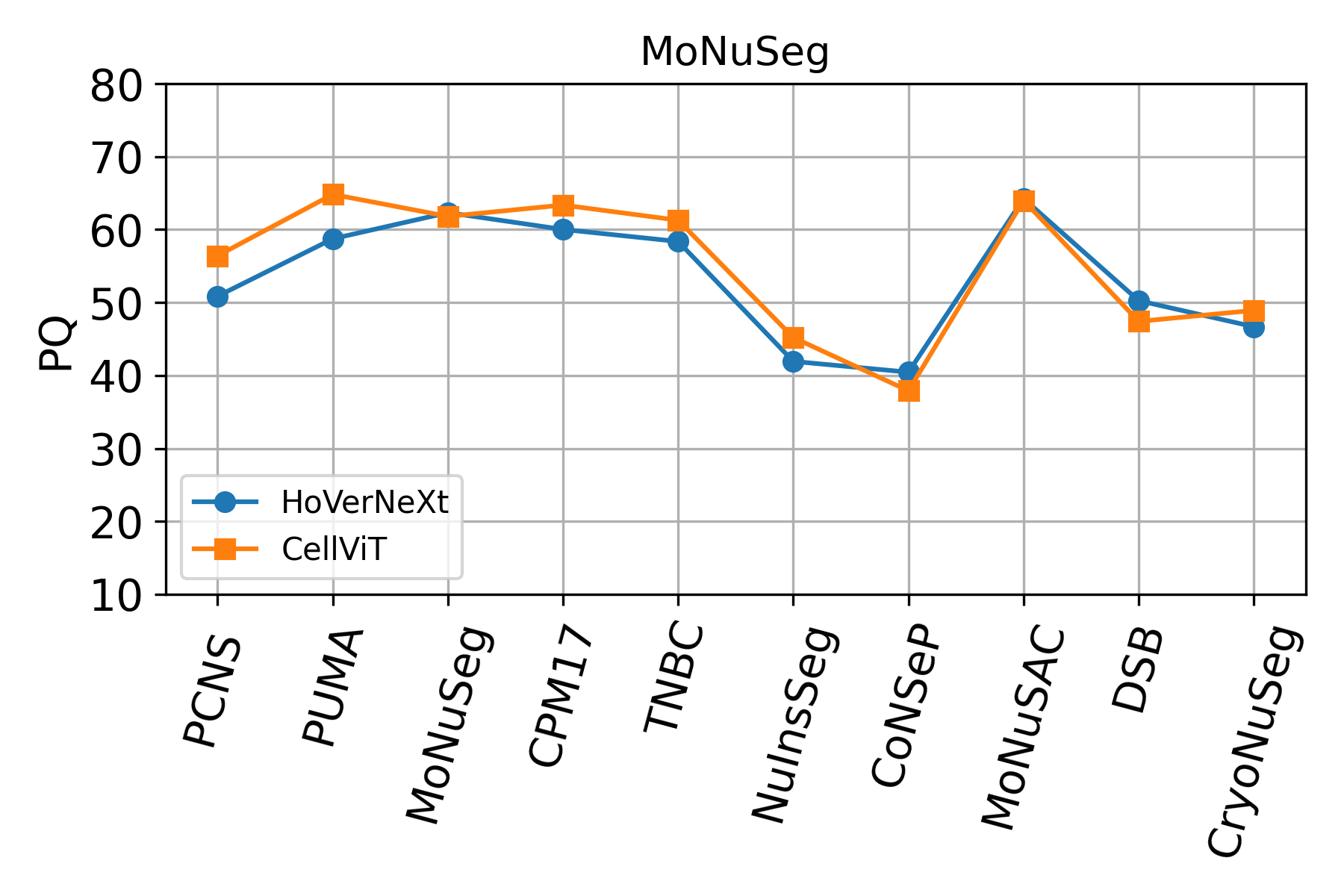}
    \end{subfigure}    
    \begin{subfigure}[b]{0.43\textwidth}
        \centering
        \includegraphics[height=3.5cm]{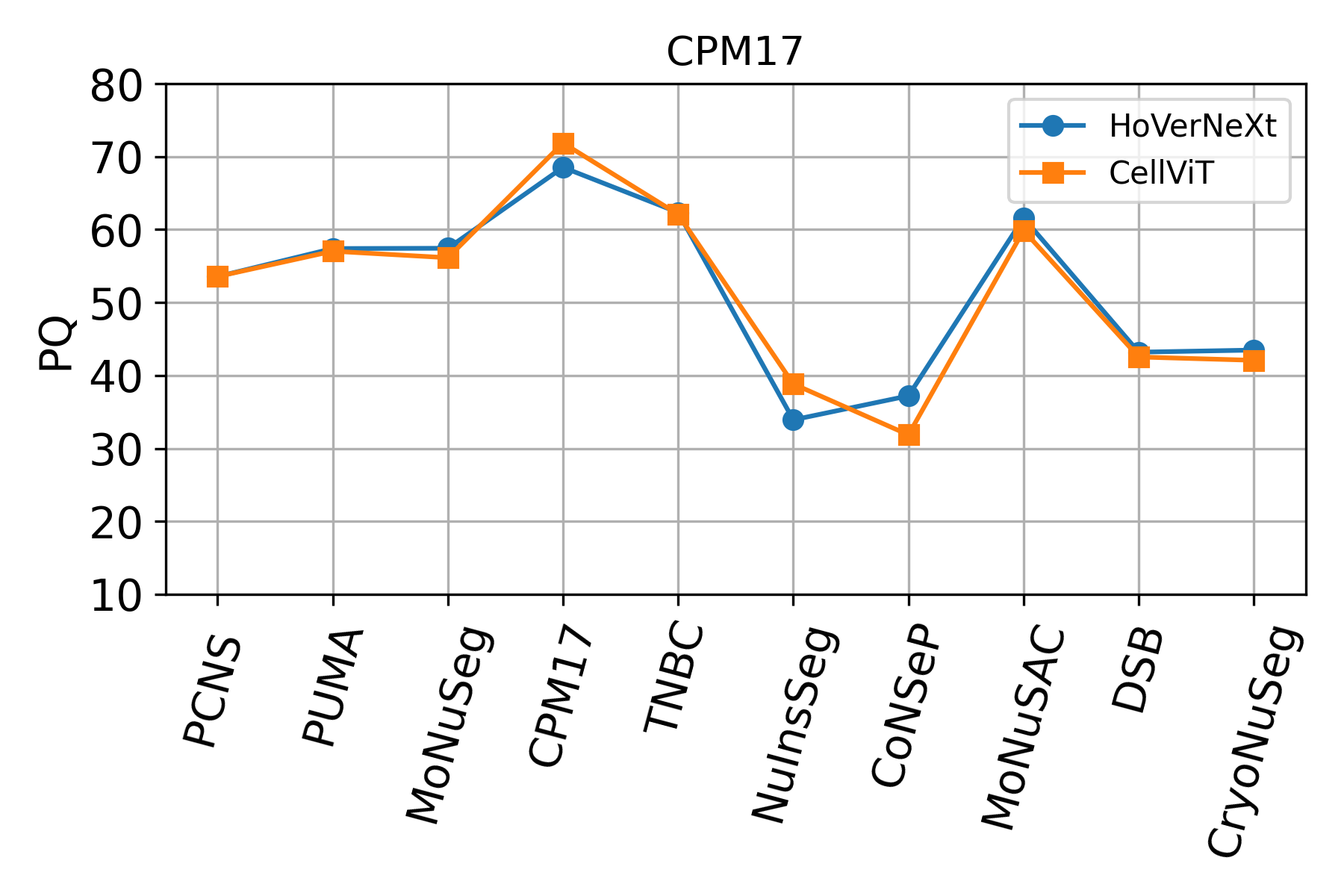}
    \end{subfigure} \\
    \begin{subfigure}[b]{0.43\textwidth}
        \centering
        \includegraphics[height=3.5cm]{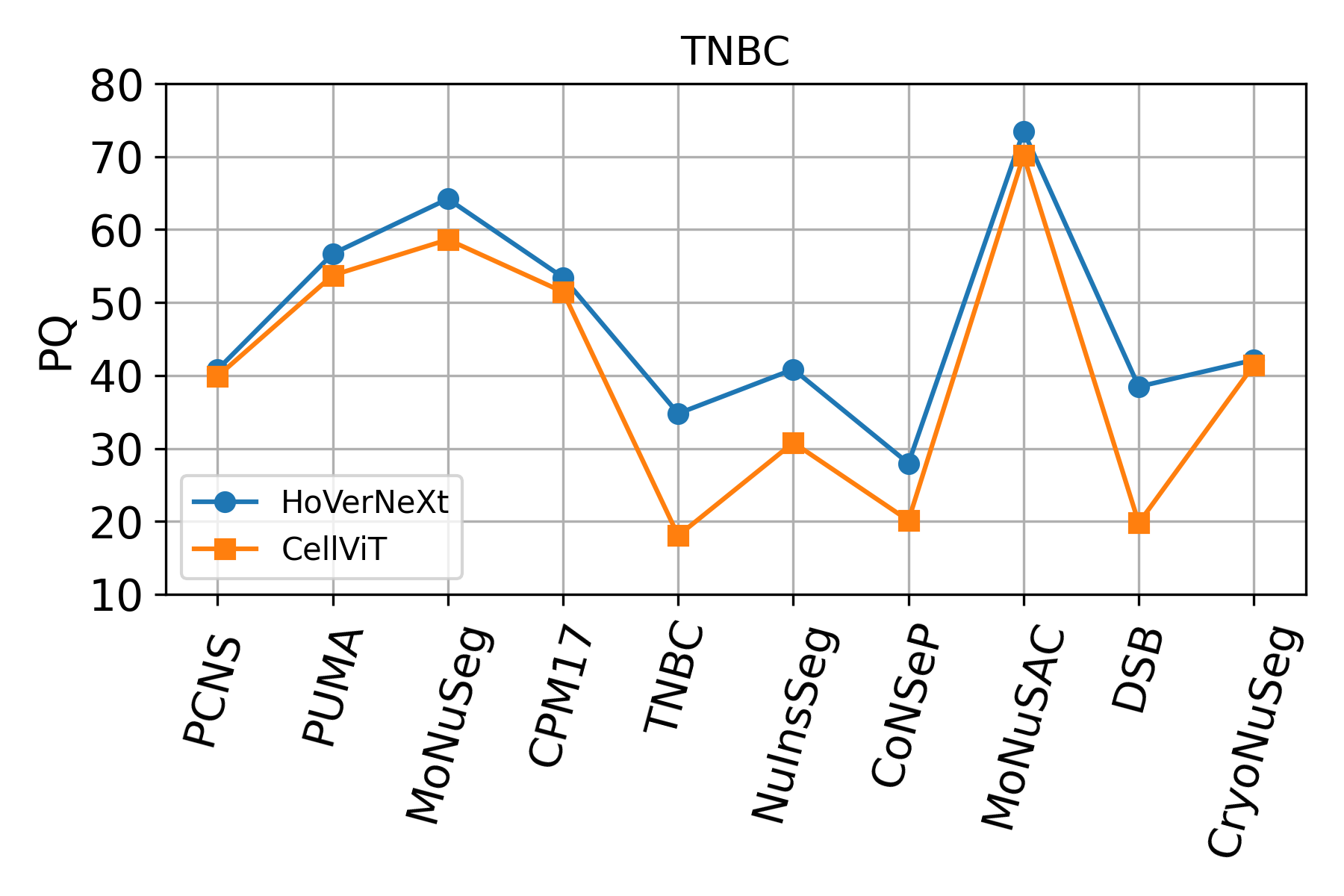}
    \end{subfigure}
    \begin{subfigure}[b]{0.43\textwidth}
        \centering
        \includegraphics[height=3.5cm]{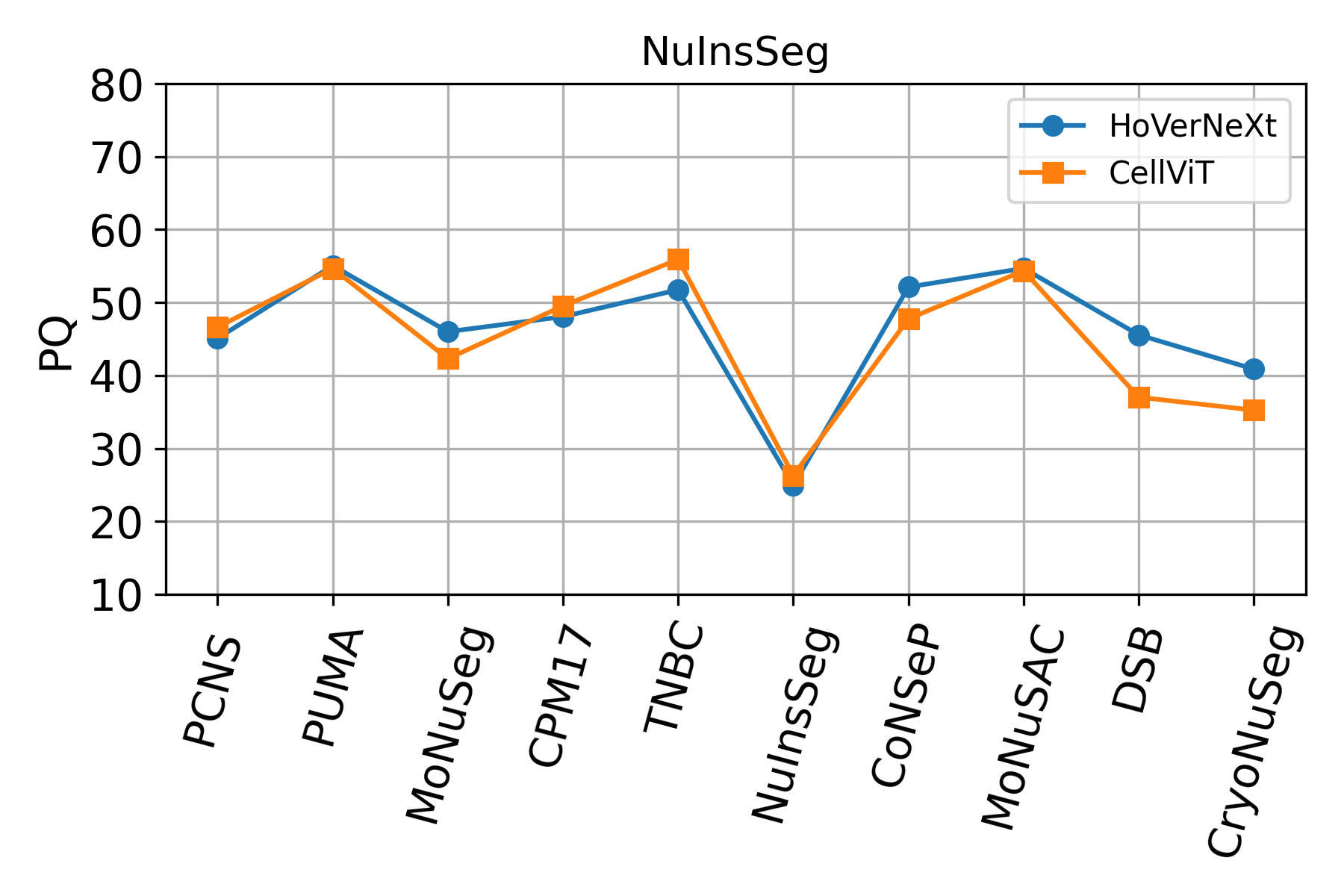}
    \end{subfigure}
        \begin{subfigure}[b]{0.43\textwidth}
        \centering
        \includegraphics[height=3.5cm]{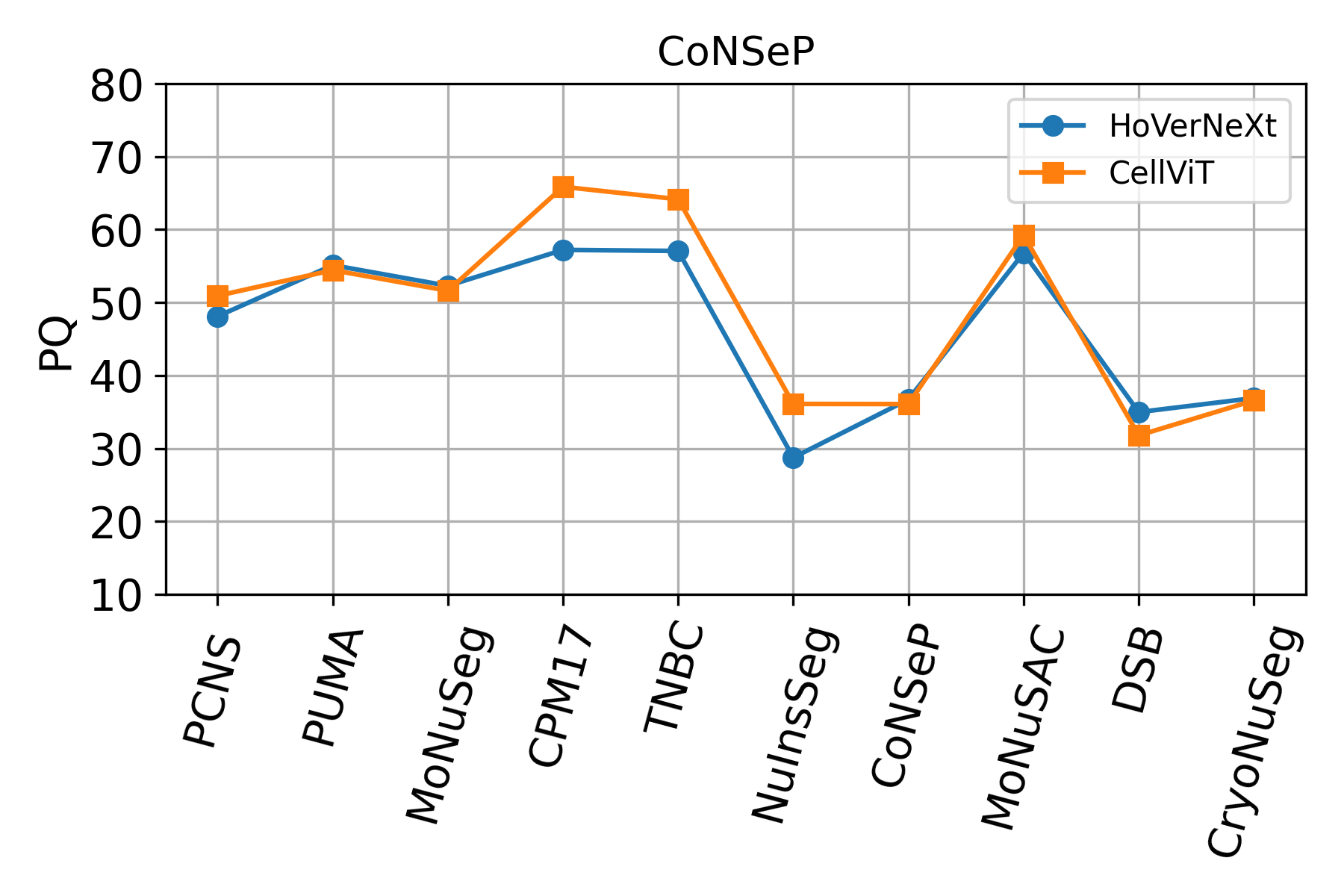}
    \end{subfigure}
    \begin{subfigure}[b]{0.43\textwidth}
        \centering
        \includegraphics[height=3.5cm]{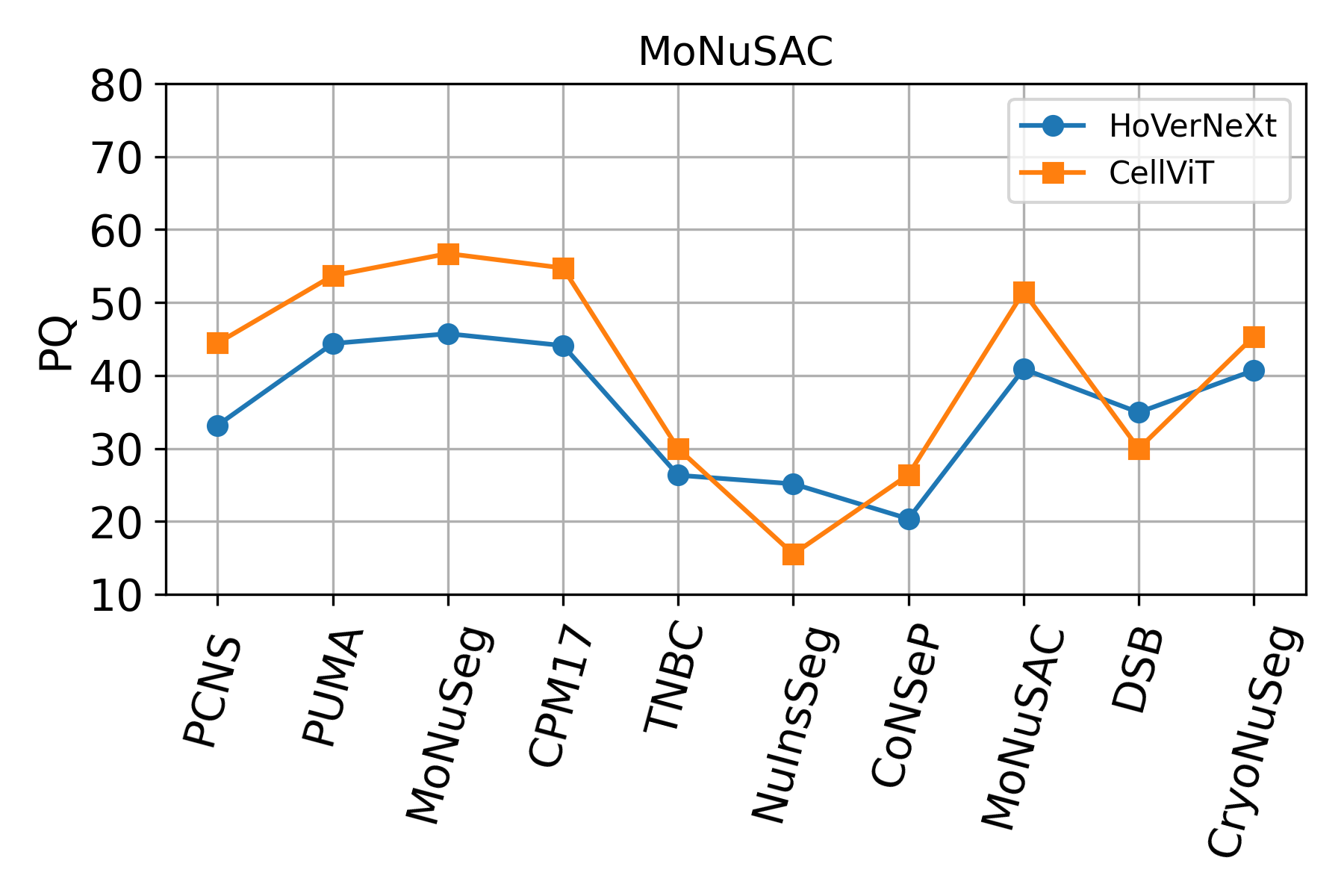}
    \end{subfigure} \\
    \begin{subfigure}[b]{0.43\textwidth}
        \centering
        \includegraphics[height=3.5cm]{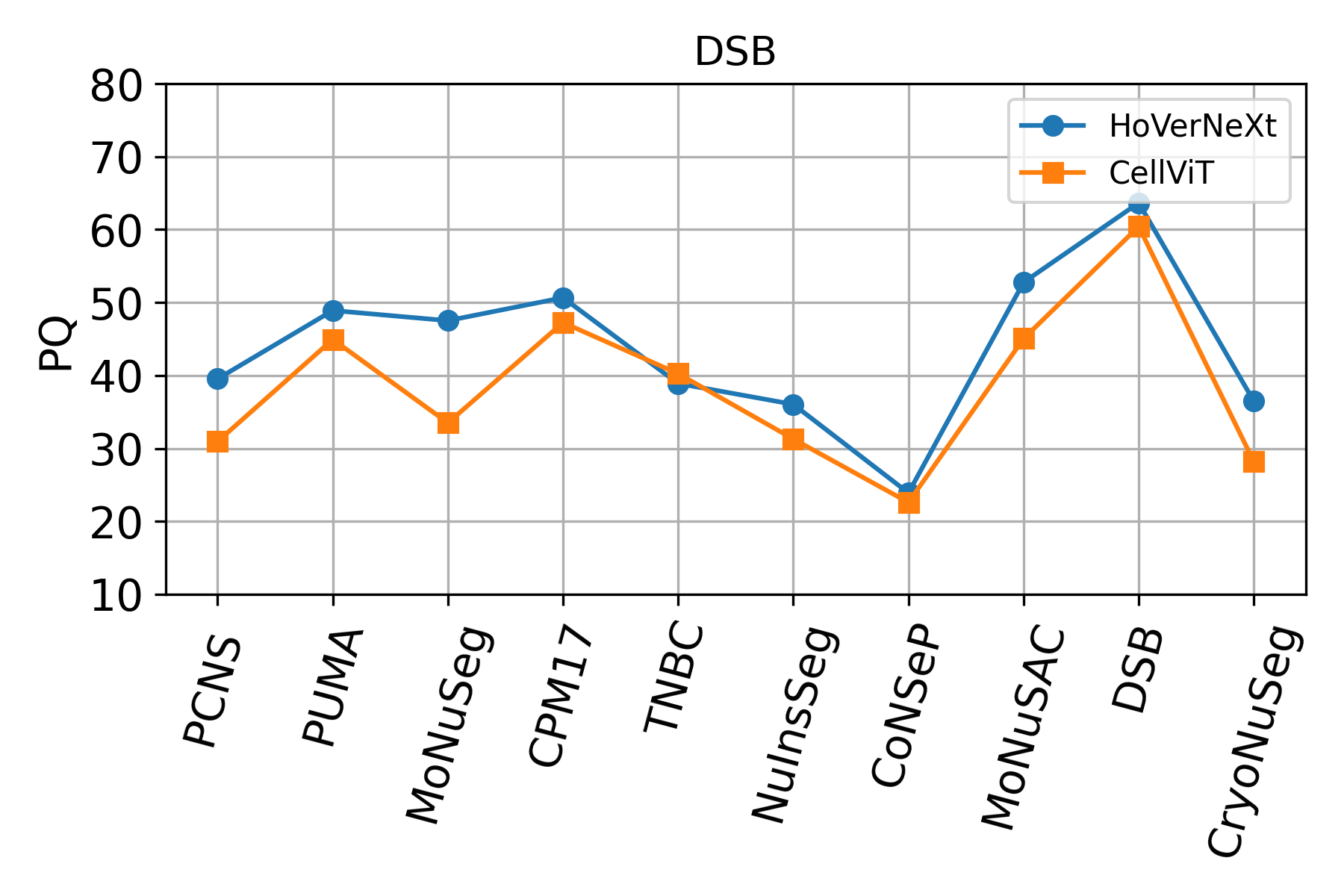}
    \end{subfigure}    
    \begin{subfigure}[b]{0.43\textwidth}
        \centering
        \includegraphics[height=3.5cm]{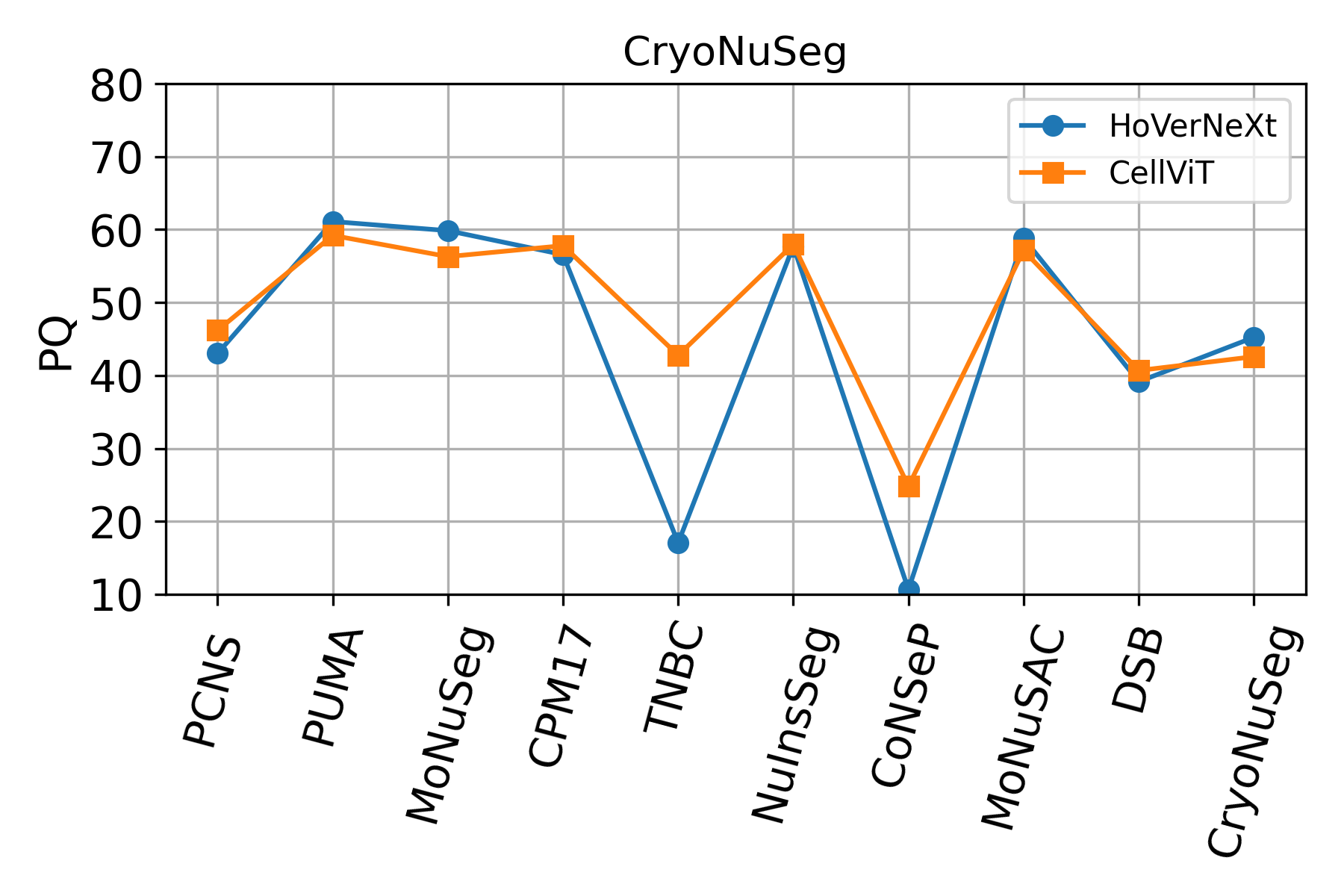}
    \end{subfigure}
    \caption{Results from Single-Dataset Training. For each training dataset, HoVerNeXt and CellViT were trained using only the corresponding dataset, and their instance segmentation performance based on panoptic quality (PQ) score was evaluated separately on each constituent subset of the unified test set (NucFuse-test).}
    \label{fig:experiment 1}
\end{figure}

\begin{figure}[]
    \centering    
    \begin{subfigure}[b]{0.48\textwidth}
        \centering
        \includegraphics[width=6cm]{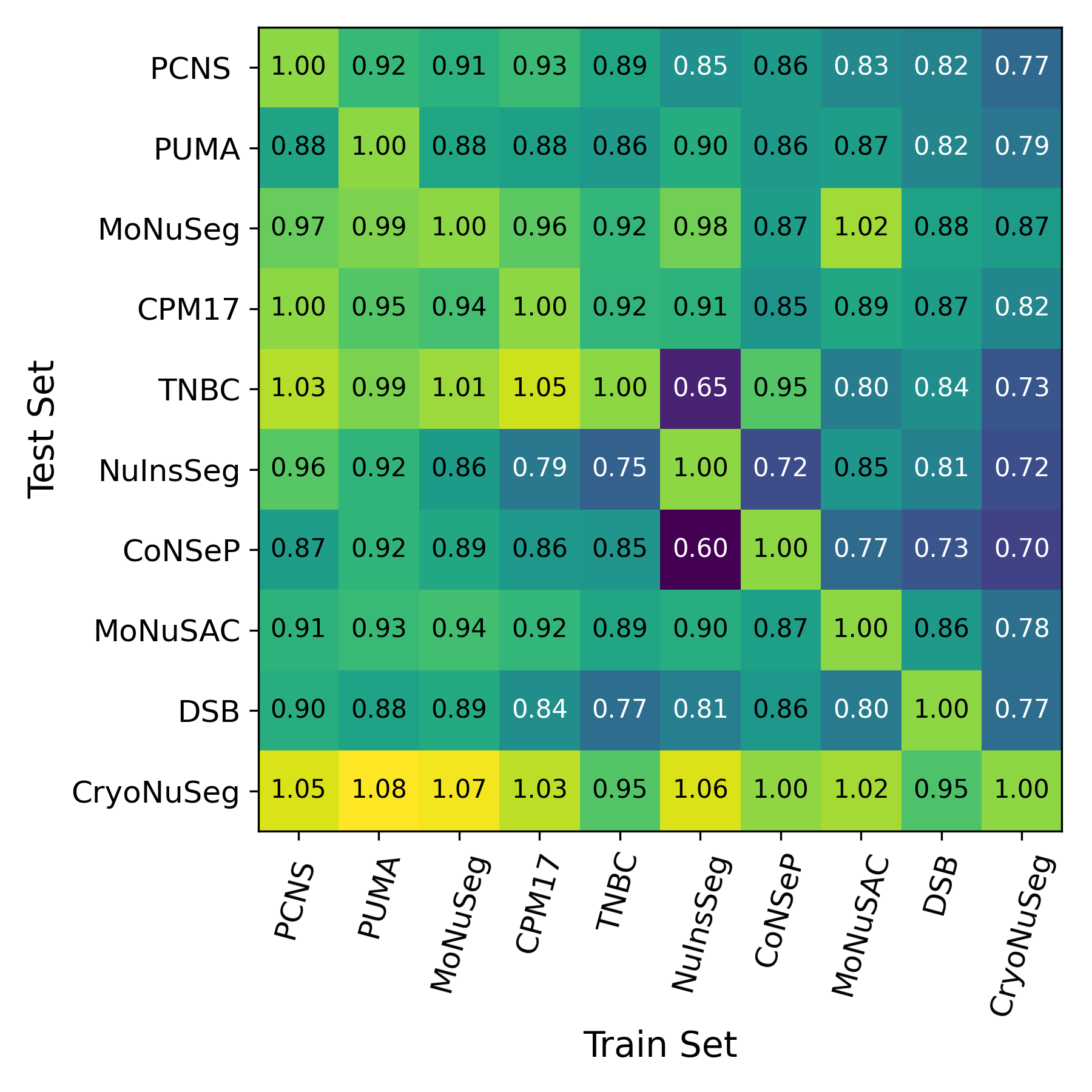}
        \footnotesize \caption{HoVerNeXt}
    \end{subfigure}
    \begin{subfigure}[b]{0.48\textwidth}
        \centering
        \includegraphics[width=6cm]{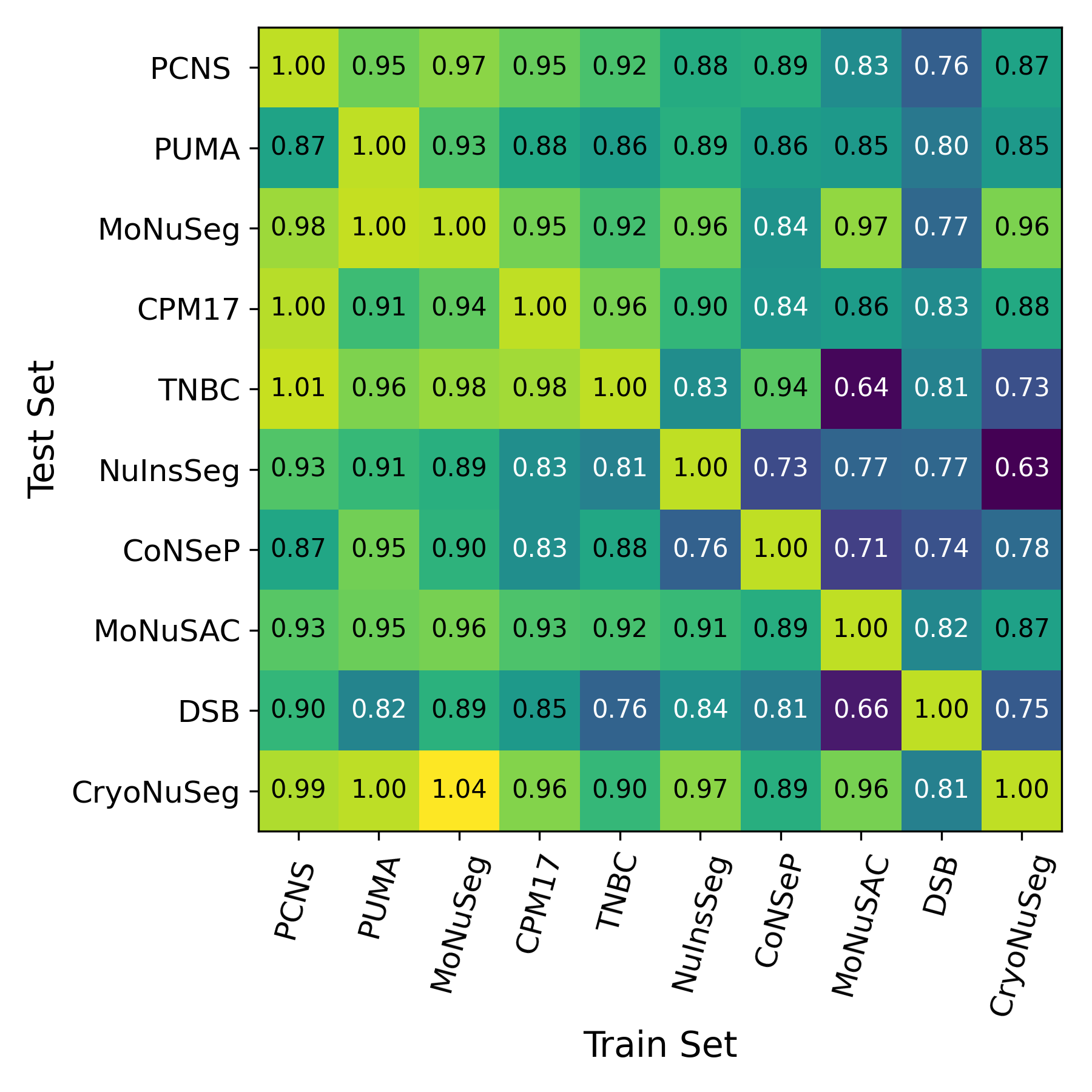}
        \footnotesize  \caption{CellViT}
    \end{subfigure}
    \caption{Dataset cross-correlation matrices for HoVerNeXt (a) and CellViT (b) models.}
    \label{fig: cross corr}
\end{figure}
\begin{figure}[]
    \centering    
    \begin{subfigure}[b]{1\textwidth}
        \centering
        \includegraphics[width=12cm]{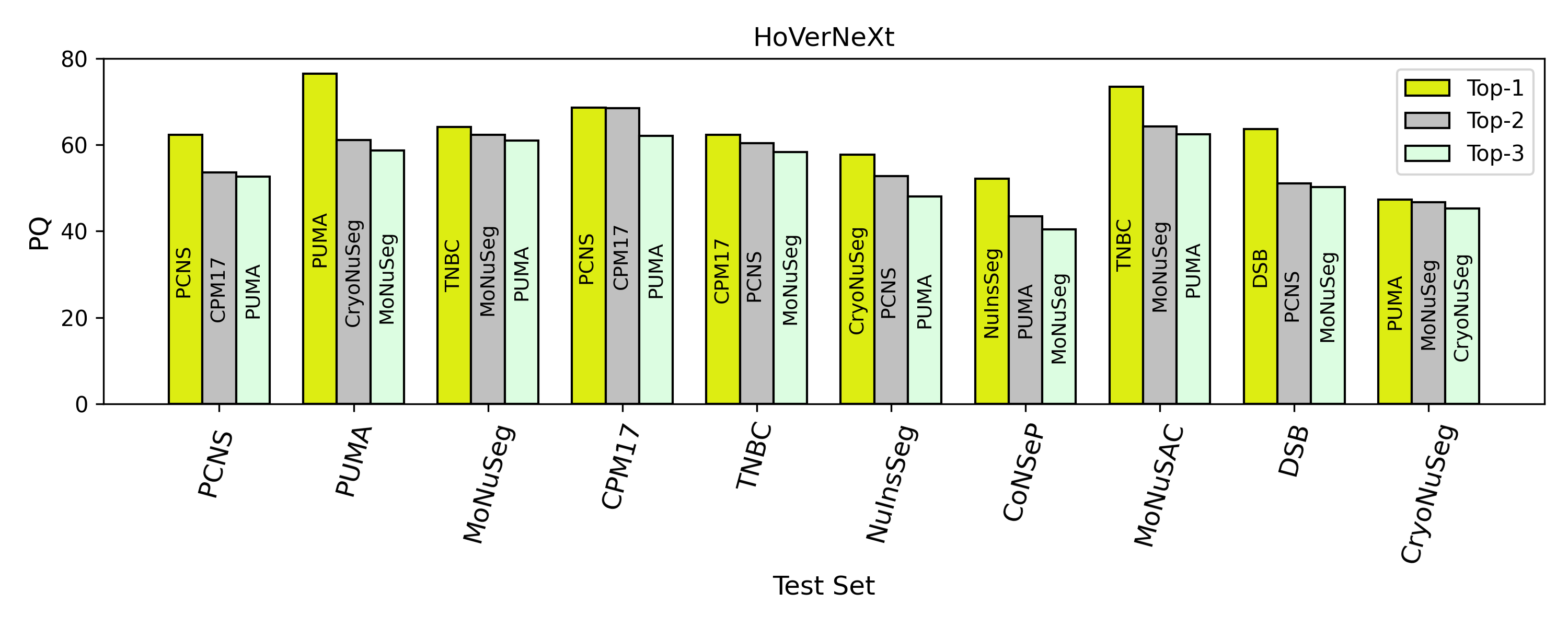}
    \end{subfigure}

    \begin{subfigure}[b]{1\textwidth}
        \centering
        \includegraphics[width=12cm]{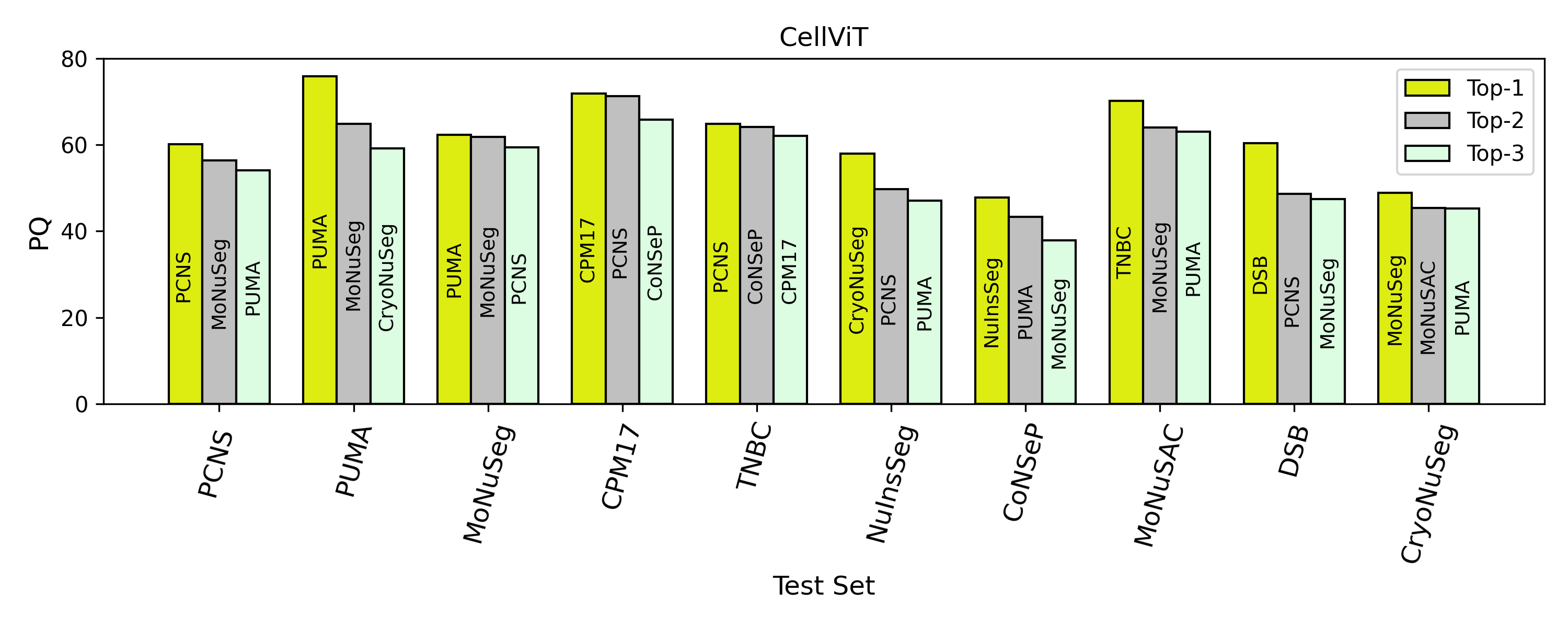}
    \end{subfigure}
    \caption{Top three performing training datasets for each test subset for HoVerNeXt (top) and CellViT (bottom).}
    \label{fig:top3}
\end{figure}
While Table~\ref{tab:single_res} reports the performance of HoVerNeXt and CellViT for each individual dataset evaluated on the entire NucFuse-test set, we provide a detailed breakdown analysis in Figure~\ref{fig:experiment 1}. This figure consists of ten subfigures, each showing performance across the different constituent subsets of the NucFuse-test set. For example, in the first subfigure (PCNS), both CellViT and HoVerNeXt are trained on the PCNS training data, and their performance is reported separately for each subset of NucFuse-test (test samples from PCNS, PUMA, MoNuSeg and others). The average performance for each test subset is reported in Table S2 in the supplementary material, where overall the models achieve the best performance on the CPM17 test subset and the worst performance on the CoNSeP test subset.

We also report the dataset-based cross-correlation results in Figure~\ref{fig: cross corr}. In each matrix, the horizontal axis represents the training dataset, while the vertical axis represents the test subset. For clearer visualization, the top three training datasets for each test subset are highlighted in Figure~\ref{fig:top3}.

From Figure~\ref{fig: cross corr}, we can observe that some values in the correlation matrices exceed 1.0. This occurs when the highest performance on a given test subset was not achieved by training on the corresponding dataset. This means training on a different dataset yielded better performance than training on the dataset’s own training split. As shown in Figure~\ref{fig:top3}, for example, for the HoVerNeXt model, the best performance on the PCNS test subset was achieved when the model was trained on the PCNS training data, which is expected. While this behavior was observed in many cases, there were also notable exceptions. For instance, on the TNBC test subset, the highest performance was obtained when the model is trained on the CPM17 training data rather than on TNBC training data. In this example, TNBC training data was not even among the top three training choices (the second and third choices were PCNS and MoNuSeg training datasets, respectively). 

Considering all values across both models, the highest cross-correlation value is 1.08 (trained on PUMA and tested on CryoNuSeg), and the lowest cross-correlation value is 0.60 (trained on NuInsSeg and tested on CoNSeP). A cross-correlation threshold can be defined to determine which datasets can be selected for training. For example, if 0.87 is chosen as the cross-correlation threshold, then for both models, the PCNS dataset alone is sufficient for training, as it achieves a cross-correlation above 0.87 across all test subsets.

\subsection{Experiment 2: K-Best Dataset Training}
\label{sec:K-Best Dataset Training}
\begin{figure}[ht]
    \centering
    \includegraphics[width=0.6\textwidth]{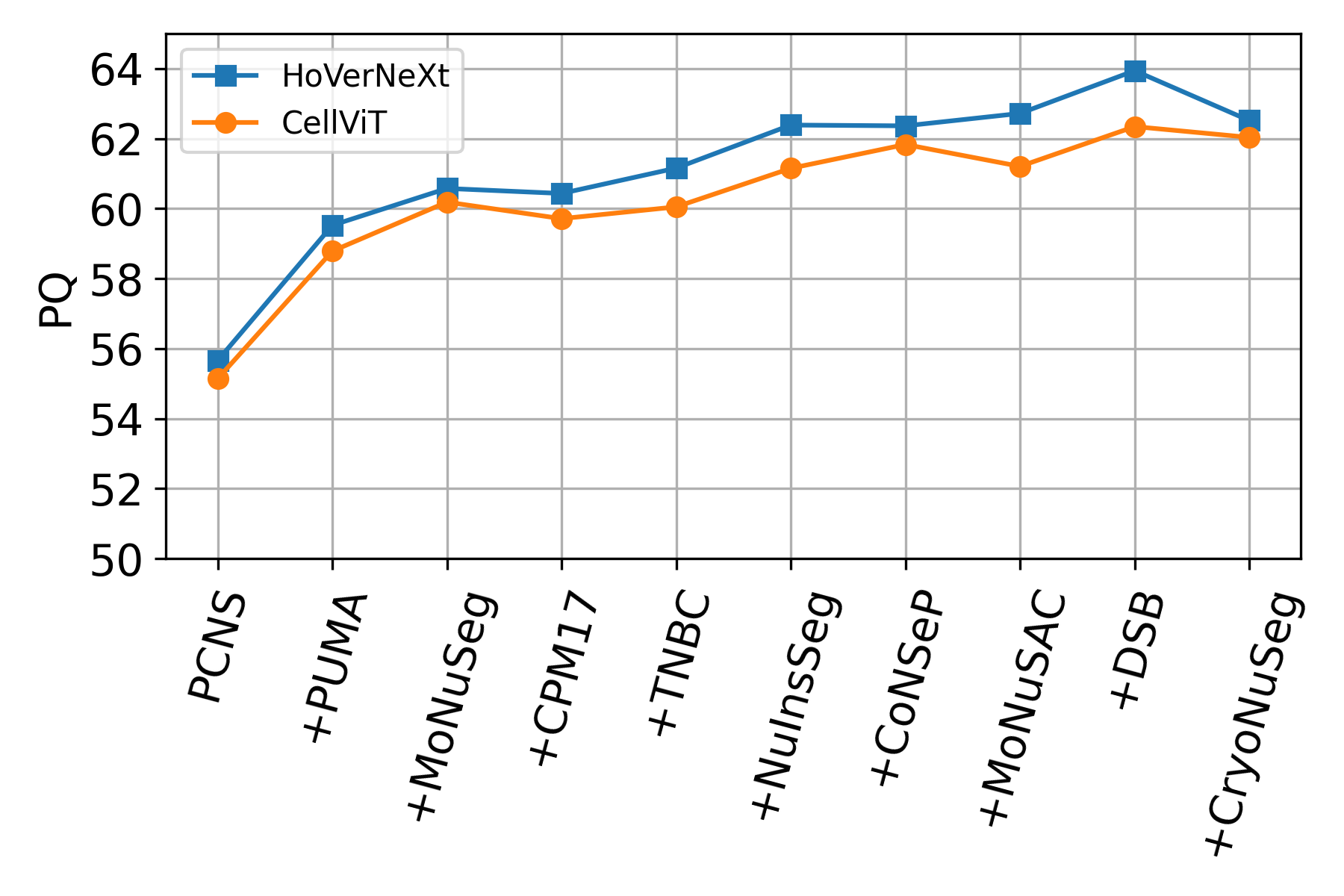}
    \footnotesize  \caption{Results of merging the top-\textit{K} best datasets and training the models, evaluated on the NucFuse-test dataset based on panoptic quality (PQ) score (\%).
    }
    \label{fig:K-Best_dataset}
\end{figure}

The results of Experiment 2, described in  Section~\ref{sec:experiment2}, are shown in Figure~\ref{fig:K-Best_dataset}. For each model, the segmentation performance on the NucFuse-test set, in terms of PQ, is reported after adding each new training dataset. The experiment included models trained with $\textit{k}=1$ (only the PCNS training data, which achieved the top overall rank in Experiment 1) up to $\textit{k}=10$, which represents training on all datasets. As can be seen for both models, performance increased with each newly added dataset up to $\textit{k}=9$. Adding the last dataset, CryoNuSeg, did not lead to improved performance. Therefore, to form the NucFuse-train dataset, we excluded CryoNuSeg and combine the training data from the remaining nine datasets (\textit{k}=9). As mentioned in Section~\ref{sec:res}, CryoNuSeg was created from frozen-sectioned samples, which differ substantially from the other datasets that are mainly based on FFPE samples. This domain shift may explain why adding CryoNuSeg to the training data is not beneficial. Comparing $\textit{k}=1$ and $\textit{k}=9$ shows an 8.28\% improvement in PQ for HoVerNeXt (from 55.66\% to 63.94\%) and a 7.20\% improvement for CellViT (from 55.14\% to 62.34\%).

Another outcome to emphasize (Figure~\ref{fig:K-Best_dataset}) is that the largest improvement in performance was achieved when the PCNS and PUMA training datasets were merged. While adding additional datasets further increased performance, their effect was less pronounced compared to the performance gain obtained by merging the first two datasets.

\begin{figure}[ht]
    \centering
    \includegraphics[width=0.6\textwidth]{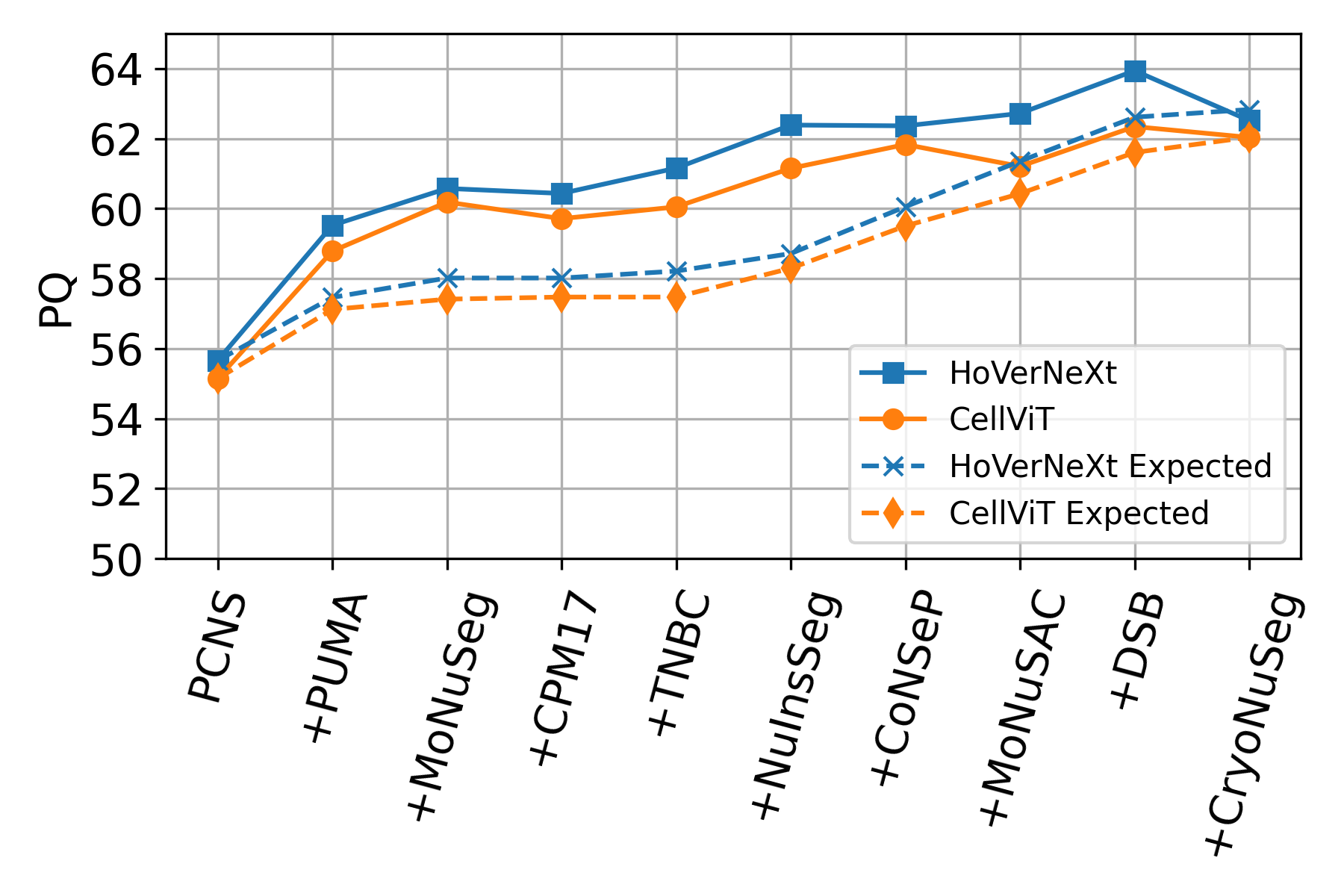}
    \footnotesize  \caption{Comparison of the expected dataset fusion improvement with the observed results in Experiment 2.  
    }
    \label{fig:expected_sum}
\end{figure}

\begin{figure}[ht]
    \centering
    \includegraphics[width=1\textwidth]{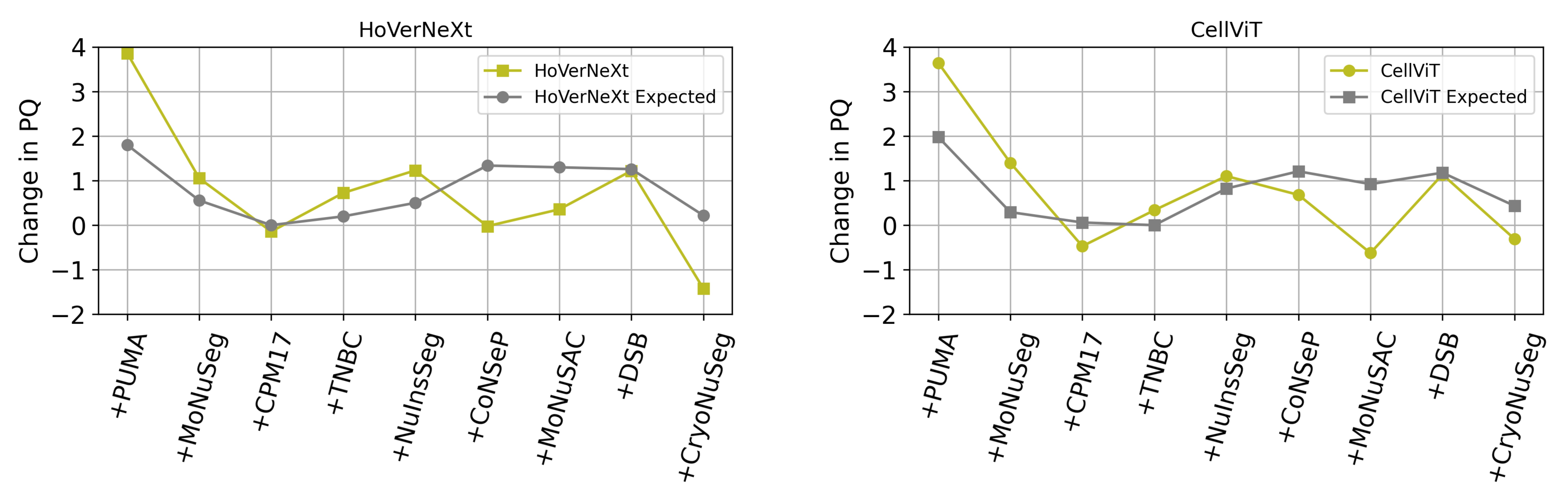}
    \footnotesize  \caption{The performance increments for both expected values and observed results for HoVerNeXt and CellViT models.
    }
    \label{fig:CellViT_hovernext_topDiff}
\end{figure}

To better understand the effect of dataset addition, we compared the expected dataset fusion improvement with the results of Experiment 2. The expected fusion improvement represents the maximum performance gain achievable by adding a new dataset for training. For each of the top-\textit{k} datasets, the expected value was calculated using the maximum PQ values from the results of Experiment 1 for the corresponding \textit{k} datasets. Figure~\ref{fig:expected_sum} presents both the expected and observed results from Experiment 2. As shown, the actual performance gain exceeds the expected improvement.
To identify which datasets contribute most to the performance gain, we computed the performance increment for both the expected values and the observed results from Experiment 2, as shown in Figure~\ref{fig:CellViT_hovernext_topDiff}. The figure reveals that some datasets produce a larger performance increase than expected (e.g., PCNS), while others contribute negatively (e.g., CryoNuSeg). A larger improvement in the observed results relative to the expected values suggests that these datasets not only enhance performance on their own samples but also positively contribute to performance on other datasets, indicating a complementary effect.

\subsection{Additional Experiments}

\subsubsection{Effect of Stain Normalization}
Within an additional experiment, we investigated the effect of stain normalization on model performance. Stain normalization is a widely used preprocessing technique in computational pathology that aims to reduce color variability caused by differences in tissue preparation, staining protocols, scanner characteristics, and laboratory-specific procedures. Such variability can introduce unwanted domain shifts that can negatively affect the generalization of DL models~\cite{HOQUE2024101997}. We investigated the effect of stain normalization using multi-target stain normalization (Avg-post technique)~\cite{ivanov2024multi} and the non-deterministic stain augmentation proposed in~\cite{mahbod2024improving}. We selected seven images from seven different organs in the MoNuSeg dataset as reference images for the non-deterministic stain augmentation during training, which was applied with a probability of 50\%~\cite{mahbod2024improving}. To conduct these experiments, a setup similar to Experiment 1 was applied, where models were trained on a single dataset and evaluated on the entire NucFuse-test set.

\begin{figure}[ht]
    \centering
    \includegraphics[width=0.6\textwidth]{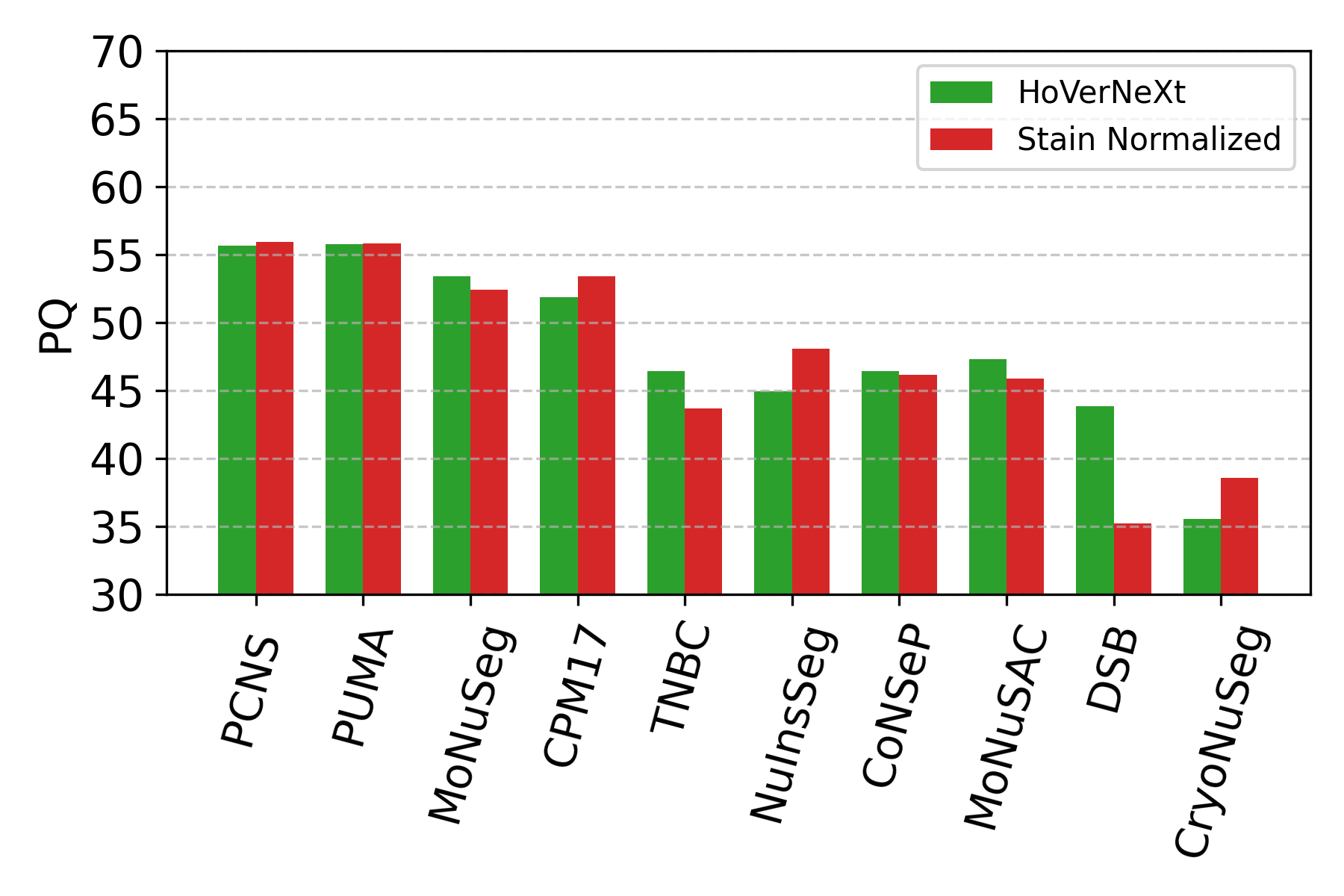}
    \footnotesize  \caption{The effect of stain normalization on the performance based on panoptic quality (PQ) score (\%). For evaluation, the entire NucFuse-test data was used. 
    }
    \label{fig:stain norm}
\end{figure}

The results of these experiments using the HoVerNeXt model are shown in Figure~\ref{fig:stain norm}. As can be seen, the effect varies across datasets. Overall, the average performance across all datasets decreased from 48.10\% to 47.49\%. Among the datasets, CPM17, NuInsSeg, and CryoNuSeg exhibited improved performance, whereas for MoNuSeg, TNBC, MoNuSAC, and DSB, performance decreased when stain normalization was applied. The most substantial decrease was observed for DSB. For PCNS, PUMA, and CoNSeP, performance remained comparable with and without stain normalization. These results confirm previous findings that stain normalization does not always improve performance in histological image analysis tasks~\cite{MOLLAHOSEYNI2025}.

\subsubsection{PanNuke and CellSAM results}
While this study focuses on publicly available, manually annotated datasets for nuclei instance segmentation, we also performed additional comparative experiments using models trained on the PanNuke dataset~\cite{10.1007/978-3-030-23937-4_2}, a semi-automatically generated dataset, and CellSAM~\cite{israel2025cellsam}, a recently developed general-purpose foundation model for cell and nuclei segmentation.

For the PanNuke experiment, we trained HoVerNeXt and CellViT using the PanNuke training data and evaluated their performance on the NucFuse-test set. We also evaluated the performance of the CellSAM model on NucFuse-test. 

The results of these experiments, together with comparisons against HoVerNeXt and CellViT trained on NucFuse-train, are reported in Table~\ref{tab:pan_nuke_pq}. As the results indicate, the models trained on PanNuke yielded inferior performance compared to those trained on NucFuse-train. This was observed despite the fact that datasets such as MoNuSeg and CPM17 were used during the creation of PanNuke, which could potentially introduce data leakage. Nevertheless, training on NucFuse-train still yielded superior performance for both HoVerNeXt and CellViT. Compared with training on individual datasets (reported in Table~\ref{tab:single_res}), PanNuke delivered superior performance; however, due to potential data leakage from MoNuSeg and CPM17, a direct comparison is not feasible. Compared with CellSAM, training on NucFuse-train achieved higher performance with a large margin. While CellSAM yielded inferior results compared to datasets in group 1 and group 2 in Table~\ref{tab:single_res}, it outperformed some datasets in group 3 and all datasets in group 4.


\begin{table}[h!]
\centering
\resizebox{\textwidth}{!}{
\begin{tabular}{lccccc}
\toprule
Dataset & \shortstack{HoVerNeXt\\(NucFuse-train)} & \shortstack{CellViT\\(NucFuse-train)} &\shortstack{HoVerNeXt\\(PanNuke)} & \shortstack{CellViT\\(PanNuke)} & CellSAM \\

\midrule
PCNS       & 61.74 & 60.05 & 62.10 & 59.16 & 40.90 \\
PUMA       & 77.09 & 75.24 & 68.58 & 67.25 & 49.88 \\
MoNuSeg    & 66.61 & 65.27 & 65.59 & 67.24 & 59.68 \\
CPM17      & 73.04 & 73.46 & 71.73 & 72.13 & 62.56 \\
TNBC       & 66.83 & 62.02 & 62.41 & 66.22 & 59.75 \\
NuInsSeg   & 59.72 & 59.12 & 48.66 & 51.54 & 46.88 \\
CoNSeP     & 50.54 & 48.64 & 50.87 & 52.34 & 30.00 \\
MoNuSAC    & 72.88 & 70.31 & 63.81 & 66.46 & 60.17 \\
DSB        & 59.00 & 60.37 & 56.94 & 38.13 & 42.31 \\
CryoNuSeg  & 51.95 & 48.95 & 50.71 & 51.26 & 35.76 \\
\midrule
\textbf{NucFuse-test} & \textbf{63.94} & \textbf{62.34} & \textbf{60.14} & \textbf{59.17} & \textbf{48.79} \\
\hline
\end{tabular}
}
\caption{Performance of HoVerNeXt and CellViT on NucFuse-test and its constituent datasets when trained on either NucFuse-train or PanNuke based on panoptic quality (PQ) score (\%). The final column reports CellSAM performance evaluated on the same test sets.}
\label{tab:pan_nuke_pq}
\end{table}

\subsubsection{Example Results}
\begin{figure}[]
    \centering    
    \includegraphics[width=1\textwidth, height=1\textheight, keepaspectratio]{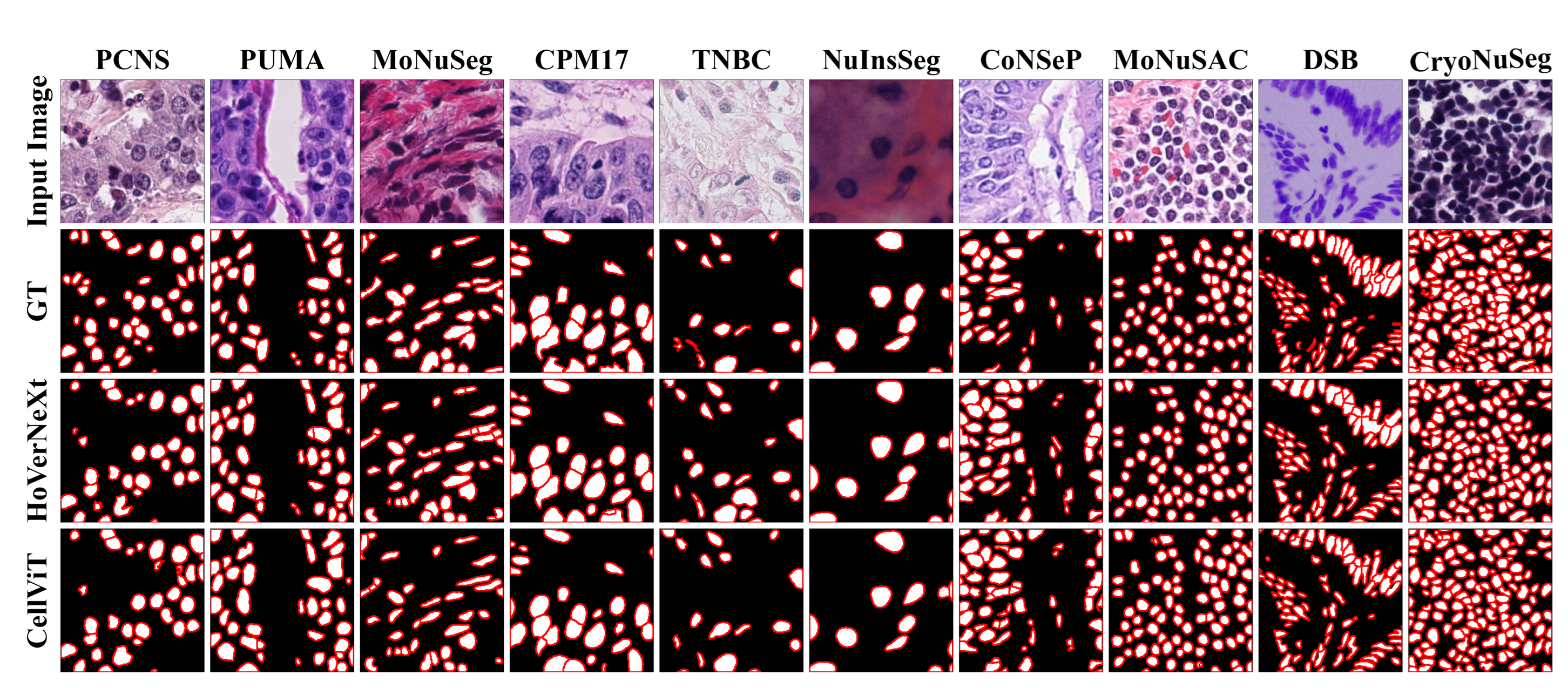}
    \caption{Visualization of nuclei instance segmentation results from  HoVerNeXt~\cite{baumann2024hover} and CellViT~\cite{HORST2024103143}. Both models were trained on NucFuse-train, and their predictions are compared with the ground truth (GT). For each image, a $256\times256$ crop is shown for better visualization. Input image: H\&E-stained images. Color code: white for nuclei, red for nuclei boundaries, and black as background.}
    \label{fig: visual1}
\end{figure}

Visualization of representative sample results is shown in Figure~\ref{fig: visual1}. The input images were randomly selected from the NucFuse-test set (one image from each test subset), and the predictions show the outputs of HoVerNeXt and CellViT when trained on the NucFuse-train dataset.

\subsection{Limitations}

Variations in dataset characteristics, such as the number of tiles, tile size, number of organs, and staining variations, make it difficult to compare datasets. In our experiments, we treated these characteristics as intrinsic properties of the datasets and designed a straightforward and unified pipeline based on this assumption.

While our approach enables fair benchmarking, improved results might be achieved for individual datasets through additional preprocessing, alternative sampling strategies during training, or by choosing a larger encoder for the models. For example, applying stain normalization improveed performance on NuInsSeg, moving it from eighth to fifth place for HoVerNeXt in Table~\ref{tab:single_res}. Another example is CryoNuSeg. CryoNuSeg (the smallest dataset in our experiments) included only 30 image tiles in total, of which 14 were used in the unified test set; therefore, only 16 images remained for training, and alternative sampling strategies could yield better results. 
However, we treated all datasets in a unified manner to enable fair comparison and to identify an optimal joint training strategy. Dataset-specific preprocessing and modifications were therefore not the main focus of this study. 

Additionally, in this study, we focused on the nuclei instance segmentation task, whereas nuclei instance segmentation combined with classification could also benefit from a similar investigation. However, considering the different nuclei classes defined across datasets, merging datasets and performing fair comparisons for the joint segmentation-and-classification task becomes more challenging. Nevertheless, this direction is valuable and remains open for future research. 




\section{Conclusion}
Dataset quality plays a critical role in training DL models, particularly for histological images characterized by complex tissue anatomy. Over the years, different research groups have introduced several publicly available datasets for nuclei instance segmentation, each with distinct properties. A key open question is determining which of these datasets exhibit greater generalizability and are therefore more suitable for benchmarking nuclei instance segmentation models. To address this, we evaluated the performance of two state-of-the-art models across multiple experimental settings and systematically compared the datasets. Based on our findings, we grouped the datasets into four categories and identified those with the highest potential for the nuclei instance segmentation task. Additionally, we proposed a unified training set and a unified test set that can serve as a benchmark for future research in nuclei instance segmentation.


\section*{Declaration of Competing Interest}
The authors declare that they have no known competing financial interests that could have appeared to influence the work reported in this paper.

\section*{Data Availability}
The datasets used in this study are publicly available and were released with the corresponding previously published papers. The code developed for this study, as well as the training and testing data (derived from the publicly available datasets listed in Table~\ref{tab:datasets}), are available on GitHub and FigShare, with the corresponding links provided in the manuscript.

\section*{Ethics Statement}
Ethical approval was not required for this study, as all datasets used are publicly available and were previously published.

\section*{Declaration of generative AI and AI-assisted technologies in the manuscript preparation process}
During the preparation of this work, the authors used ChatGPT (version 5.2) to check grammar and spelling and, at times, improve the readability of some sentences. After using these tools, the authors reviewed and edited the content as needed and take full responsibility for the content of the publication.

\section*{CRediT Authorship Contribution Statement}
\textbf{Nima Torbati:} Conceptualization, Methodology, Writing – original draft, Writing – review and editing, Formal analysis, Data curation, Validation, Software. 
\textbf{Anastasia Meshcheryakova:} Supervision, Writing – review and editing, Visualization, Funding acquisition, Project administration.
\textbf{Ramona Woitek:} Supervision, Writing – review and editing, Project administration.
\textbf{Sepideh Hatamikia:} Supervision, Writing – review and editing, Project administration. 
\textbf{Diana Mechtcheriakova:} Supervision, Writing – review and editing, Visualization, Funding acquisition, Project administration.
\textbf{Amirreza Mahbod:} Supervision, Conceptualization, Methodology, Writing – original draft, Writing – review and editing, Funding acquisition, Formal analysis, Data curation, Validation, Project administration. 

\section*{Acknowledgment}
This work was supported by the Vienna Science and Technology Fund (WWTF) and by the State of Lower Austria [Grant ID: 10.47379/LS23006].

\bibliographystyle{elsarticle-num} 
\bibliography{refs.bib}

\end{document}